\documentclass[journal,twoside,web]{ieeecolor}
\usepackage{tmi}
\usepackage{cite}
\usepackage{amsmath,amssymb,amsfonts}
\usepackage{multirow}
\usepackage{multicol}
\usepackage{algorithmic}
\usepackage{graphicx}
\usepackage{adjustbox}
\usepackage{booktabs}
\usepackage{textcomp}
\usepackage{xcolor}
\usepackage{makecell}
\definecolor{bb}{rgb}{0.0, 0.0, 0.5}
\definecolor{Gray}{gray}{0.9}
\usepackage{colortbl}
\usepackage{bbding}
\usepackage[colorlinks,linkcolor=red]{hyperref}
\def\etal{\emph{et al.}}

\newcommand{\eg}{\textit{e.g.}}
\newcommand{\Amendment}[1]{{#1}}
\newcommand{\Amendmentsecond}[1]{{#1}}
\newcommand{\Amendmentthird}[1]{{#1}}

\def\BibTeX{{\rm B\kern-.05em{\sc i\kern-.025em b}\kern-.08em
    T\kern-.1667em\lower.7ex\hbox{E}\kern-.125emX}}
\begin{document}
\title{Digital Staining with Knowledge Distillation: A Unified Framework for Unpaired and Paired-But-Misaligned Data}
\author{Ziwang Xu, Lanqing Guo, Satoshi Tsutsui, Shuyan Zhang, Alex C. Kot, \IEEEmembership{Life Fellow, IEEE} and  \\Bihan Wen, \IEEEmembership{Senior Member, IEEE}
\thanks{This work was partially supported by the National Research Foundation Singapore Competitive Research Program (award number CRP29-2022-0003), and by the Agency for Science, Technology and Research, Singapore, under Grant CDA 202D800042. \textit{(Corresponding author: Bihan Wen.)}}
\thanks{Ziwang Xu, Satoshi Tsutsui, Alex C. Kot, and Bihan Wen are with the Rapid-Rich Object Search Lab, Nanyang Technological University 
(email: \{ziwang001, lanqing001\}@e.ntu.edu.sg; \{satoshi.tsutsui, eackot, bihan.wen\}@ntu.edu.sg).
}
\thanks{Lanqing Guo currently is a postdoc research fellow at The University of Texas at Austin (email: lanqing.guo@austin.utexas.edu)}
\thanks{Shuyan Zhang was in the Institute of Materials Research and Engineering, Agency for Science, Technology and Research (A*STAR) (email: shuyanzhang@alumni.harvard.edu).}
}

\maketitle

\begin{abstract}
Staining is essential in cell imaging and medical diagnostics but poses significant challenges, including high cost, time consumption, labor intensity, and irreversible tissue alterations.
Recent advances in deep learning have enabled digital staining through supervised model training. 
However, collecting large-scale, perfectly aligned pairs of stained and unstained images remains difficult.
In this work, we propose a novel unsupervised deep learning framework for digital cell staining that reduces the need for extensive paired data using knowledge distillation. 
We explore two training schemes: (1) unpaired and (2) paired-but-misaligned settings.
For the unpaired case, we introduce a two-stage pipeline, comprising light enhancement followed by colorization, as a teacher model.
Subsequently, we obtain a student staining generator through knowledge distillation with hybrid non-reference losses. 
To leverage the pixel-wise information between adjacent sections, we further extend to the paired-but-misaligned setting, adding the Learning to Align module to utilize pixel-level information. Experiment results on our dataset demonstrate that our proposed unsupervised deep staining method can generate stained images with more accurate positions and shapes of the cell targets in both settings. \Amendmentthird{Compared with competing methods, our method achieves improved results both qualitatively and quantitatively (e.g., NIQE and PSNR).}  \Amendment{We applied our digital staining method to the White Blood Cell (WBC) dataset, investigating its potential for medical applications.} 
The dataset and code will be released at \href{https://github.com/wwxb2012/digital_staining_knowledge_distillation}{link}.
\end{abstract}


\section{Introduction}
\label{sec:introduction}

\IEEEPARstart{S}{taining} is a pivotal procedure in cell imaging and medical diagnosis, enabling clinicians to assess both morphological and chemical information within a microscopic context through enhanced imaging contrast.
Hematoxylin and Eosin (H\&E) stain is the most widely employed dye staining technique~\cite{feldman2014tissue}.

However, the H\&E staining process is considerably time-consuming and expensive in practical applications. \Amendment{Specifically, it requires approximately 45 minutes to complete and incurs a cost ranging from  \$$2$ to \$$5$ per} \Amendment{slide~\cite{rivenson2019phasestain}}. 
Moreover, despite standardized protocols, the outcomes of cell staining often exhibit variations across distinct histopathology laboratories due to the specific staining conditions, leading to domain shifts~\cite{cheng2023disentangled} that negatively impact on subsequent diagnostic procedures. Given the irreversibility of the staining process, unsatisfactory results require preparing new cell samples and conducting the entire staining process again.
\Amendmentsecond{In summary, the H\&E staining process is time-consuming, costly, prone to variability across labs, and requiring repeated procedures due to its irreversibility when results are unsatisfactory.}

\begin{figure}[!t] 
\centering 
\includegraphics[width=0.5\textwidth]{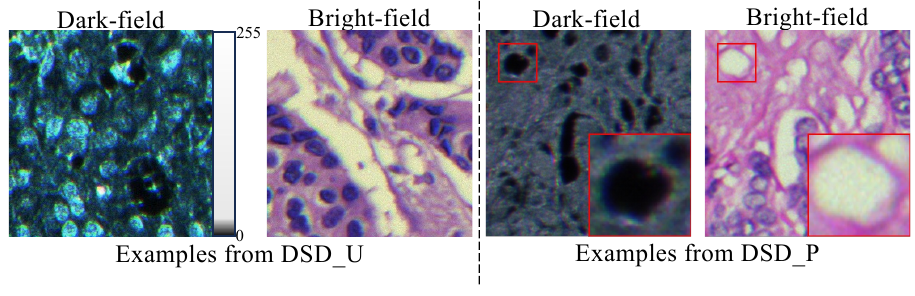} 
\caption{Examples of our unpaired dataset (DSD\_U) and paired-but-misaligned dataset (DSD\_P). For two images from the DSD\_P dataset, by comparing the black part in the DF image and the white part in the BF image, we can see that cavity areas dilate obviously, which contributes to misalignment. \Amendment{ We enhanced the dark-field images from DSD\_U and added a color bar for improved visualization, the same for other examples of dark-field images from DSD\_U in this paper.} } 
\label{fig:dataset_samples} 
\vspace{-5mm}
\end{figure}
To address these constraints associated with conventional physical staining, recent studies have introduced deep learning-based approaches for digital staining in microscopy cell images acquired through various modalities, such as Quantitative Phase Imaging (QPI)~\cite{rivenson2019phasestain}, Autofluorescence Imaging~\cite{zhang2020digital,rivenson2019virtual,rivenson2020emerging,fang2023virtual,doanngan2022label}, Whole Slide Imaging (WSI)~\cite{rana2020use}, Ultraviolet photoacoustic microscopy~\cite{kang2022deep,cao2023label} and Differential Interference Contrast (DIC) microscopy~\cite{tomczak2023digital}.
\Amendment{Dark-field imaging~\cite{verebes2013hyperspectral,chua2022label,zhang2023label} is well suited for this application because it rejects the unscattered background light from the image and produces excellent imaging contrast for transparent and thin samples. In addition, dark-field imaging is relatively easier to implement and more cost-effective than other optical contrast imaging techniques.}
\Amendmentsecond{Throughout this paper, we refer to the transition from unstained to stained images as digital staining, and the} 
\Amendmentsecond{transition from one type of staining to another as virtual staining.}

Additionally, prevailing deep learning techniques for digital staining primarily adopt a supervised approach~\cite{rivenson2019phasestain,zhang2020digital,rivenson2019virtual,rivenson2020emerging,rana2020use}. These methods necessitate extensive datasets comprising meticulously aligned pairs of unstained and stained cell images for training the models. 
However, in practical scenarios, acquiring such datasets poses substantial challenges due to the following reasons: (1) Distortions in tissue components during the processing procedure make precise pairing between bright-field and dark-field images unattainable. (2) Ensuring consistent slide orientation across diverse imaging trials is challenging, resulting in rotational and translational discrepancies in staining image pairs even when originating from the same slide.    Consequently, the inaccurate alignment between training pairs leads to pronounced performance degradation using fully-supervised methods, which introduces errors in subsequent medical diagnoses.

\Amendmentsecond{
\Amendmentthird{To address the challenges of collecting the paired and fully aligned datasets,}
we propose an effective approach for unsupervised digital staining of dark-field images~\cite{xu2023unsupervised}, utilizing a Generative Adversarial Network (GAN) for unpaired datasets. Our approach relies on a teacher-student paradigm, where the teacher model generates intermediate references through light enhancement and colorization, guiding the student model to learn the staining process in an end-to-end manner through knowledge distillation. }

\Amendmentsecond{The teacher model breaks the staining process into two stages: light enhancement and colorization. This ensures structural consistency in the intermediate outputs. While the teacher model predicts images with structural information, the level of detail remains insufficient. To address this, we introduce}
\Amendmentsecond{a novel knowledge distillation process to develop a student Generative Adversarial Network (GAN) that incorporates hybrid non-reference loss functions. In the student model, the GAN facilitates the simultaneous transfer of the style from real stained images, while the knowledge distillation loss preserves both style and structure in the absence of alignment.}

\Amendmentthird{
While the above approach is effective for unpaired settings, we encounter paired settings, which provide additional information to further enhance staining performance. Specifically, we can obtain pairs of stained and unstained images by capturing unstained images and stained images from adjacent sections. Although these images are spatially paired, they are not aligned at the pixel level. For such paired-but-misaligned datasets, where partial pixel-level correspondence is available, we introduce the Learning to Align (LtA) module in {\color{cyan}Section~\ref{sec:registration_part}} in the paired-but-misaligned setting. This module learns geometric transformations to align misaligned pairs, enabling the framework to effectively use additional information from adjacent sections while preserving structural details. By incorporating this module, our method further resolves alignment challenges specific to paired settings.}

\Amendmentthird{
Our framework addresses two distinct but related settings in digital staining: (1) the problem of unpaired digital staining (where no pixel-level correspondence exists between unstained and stained images), with our method in {\color{cyan}Section~\ref{sec:teacher_model}} and {\color{cyan}Section~\ref{sec:distillation}}; and (2) the paired-but-misaligned setting, }
\Amendmentthird{addressed in {\color{cyan}Section~\ref{sec:registration_part}}, involves an alignment challenge that arises when unstained and stained images are derived from adjacent tissue sections.}
\Amendmentthird{While adjacent sections share structural content, they exhibit geometric mismatches at the pixel level.
}

Additionally, to establish a benchmark for evaluating diverse digital staining techniques applied to dark-field microscopy cell images, we introduce the novel \textbf{D}igital \textbf{S}taining \textbf{D}ataset (\textbf{DSD}) comprising authentic cell images. The dataset comprises distinct sets of both unstained dark-field images and stained bright-field images. These datasets are collected from the unpaired setting or the paired-but-misaligned setting, named as DSD\_U dataset and DSD\_P dataset respectively. A selection of examples from our dataset, from both DSD\_U and DSD\_P, is depicted in {\color{cyan}Figure~\ref{fig:dataset_samples}}.

\begin{figure}[t] 
\centering 
\includegraphics[width=0.4\textwidth]{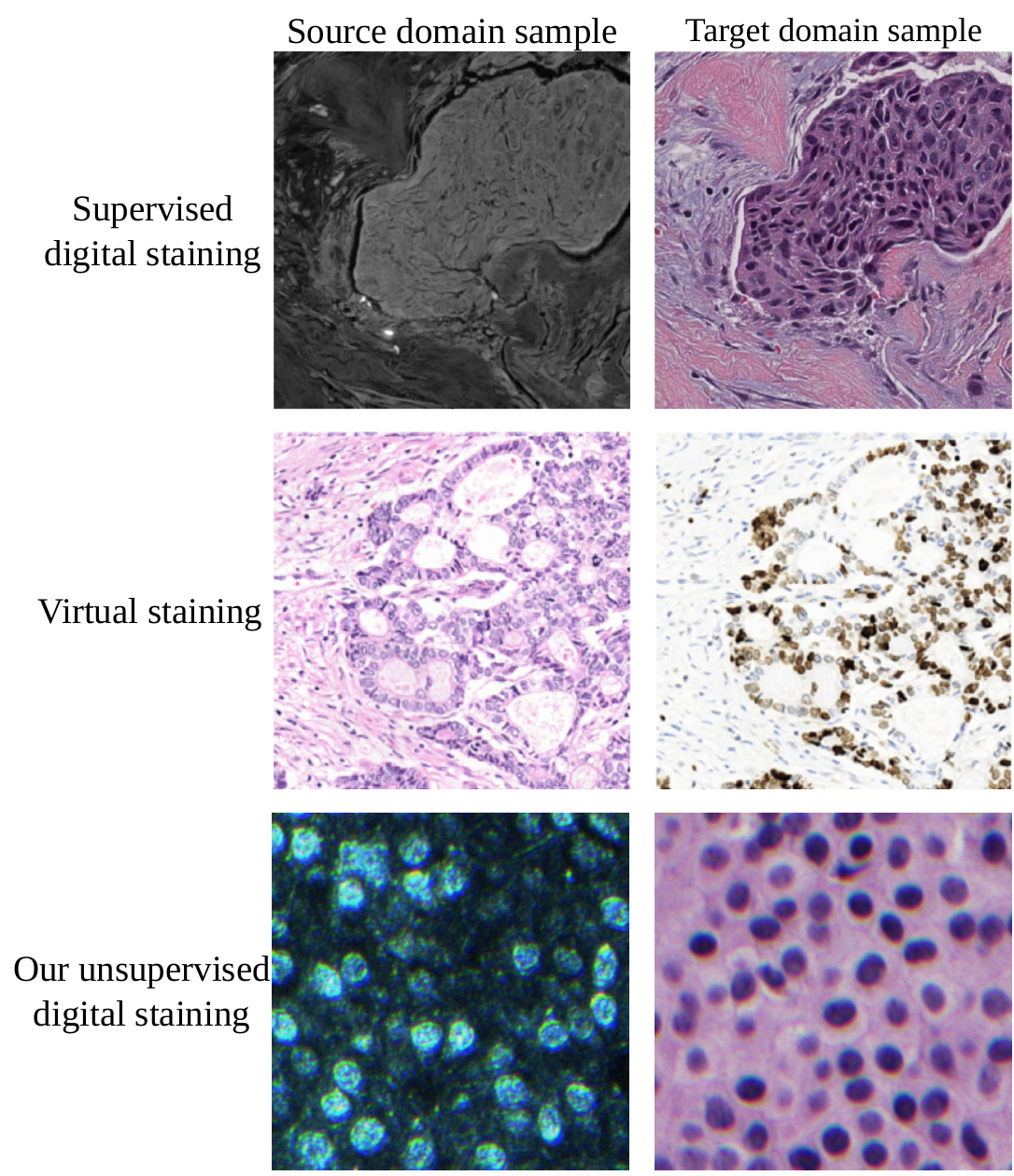} 
\caption{Illustrations of different digital staining categories. The samples for supervised digital staining are from~\cite{rivenson2019virtual}, and the samples for virtual staining are from~\cite{liu2021unpaired}. \Amendment{We uniformly refer to the transition from unstained to stained as \textbf{``digital staining"} and the transition from one type of staining to another as \textbf{``virtual staining"} throughout this paper.}} 
\label{fig:staining_category} 
\vspace{-3mm}
\end{figure}
Experimental findings corroborate the efficacy of our approach, demonstrating its notable performance improvements over the suggested staining dataset in comparison to prevalent unsupervised deep generative models. 
\Amendment{We also apply our digital staining method to the White Blood Cell (WBC) dataset to investigate potential medical applications.}

\begin{table*}[t]
\vspace{-4mm}
\centering
\footnotesize
\renewcommand{\arraystretch}{0.8}
\caption{Summary of recent deep digital staining \& virtual staining works. \textbf{P} and \textbf{U} denote \textbf{P}aired and \textbf{U}npaired supervision in the training stage respectively. }
\adjustbox{width=.95\linewidth}{
    \begin{tabular}{cccccc}
        \toprule
         Method & Original modality & Target modality & Tissue & \textbf{P}/\textbf{U} & \textbf{Digital}/\textbf{Virtual} Staining  \\\midrule
         \cite{salido2023comparison} & Unstained multispectral  & H\&E & Breast & P+U & Digital  \\
         \cite{rivenson2019phasestain} & Quantitative phase image & H\&E, Jones’ stain, Masson’s trichrome &Skin, Kidney, Liver & P & Digital \\
         \cite{zhang2020digital} & Autofluorescence & H\&E, Jones’ stain, Masson’s trichrome & Kidney & P & Digital  \\
         \cite{rivenson2019virtual}& Autofluorescence & H\&E, Jones’ stain, Masson’s trichrome &
         \makecell[c]{Salivary gland, Thyroid,\\ Kidney, Liver, Lung} & P & Digital\\
         \cite{rivenson2020emerging} & Autofluorescence & H\&E, Jones’ stain, Masson’s trichrome & Liver, Kidney & P & Digital  \\
         \cite{rana2020use} & Whole slide image &  H\&E& Prostate& P & Digital  \\
         
         \cite{fang2023virtual}  & Autofluorescence & H\&E & Lymphoid & P & Digital\\
          \cite{doanngan2022label} & Autofluorescence & HER2 & Breast & P & Digital \\
         \cite{UTOM-li2021unsupervised} & Autofluorescence & H\&E & colon & U & Digital  \\
         \cite{pillar2022virtual} & Autofluorescence & H\&E, Jones’ stain, Masson’s trichrome & \XSolidBrush  & \XSolidBrush  & Digital  \\
         \cite{kang2022deep} & Ultraviolet photoacoustic image & H\&E & Mouse brain & U & Digital  \\
         \cite{cao2023label} & Ultraviolet photoacoustic image & H\&E & Bone & U & Digital \\
         \cite{pradhan2021computational} & Non-Linear Multimodal (NLM) image & H\&E & \XSolidBrush &P+U & Digital\\
         \cite{liu2021unpaired} & H\&E & Ki-67 & Breast, Neuroendocrine & U & Virtual \\
         \cite{lahiani2019virtualization} &Ki67-CD8 & FAP-CK & Liver  & U & Virtual \\
         \cite{PEC-lahiani2020seamless} & H\&E & FAP-CK & Liver & U & Virtual \\
         \cite{sun2023bi}  & H\&E/CK7 & CK7/H\&E & Breast, Lung& U & Virtual\\
         \cite{URUST-ho2022ultra} & H\&E & CD4, CD8, Ki-67, PR, EGFR & breast, Lung, Coad &U& Virtual \\
         
         \cite{biswas2023generative} &  
         Hyperspectral H\&E & EVG & Pancreas & U & Virtual\\
         \cite{de2021deep} &  H\&E & PAS, Jones’ stain, Masson’s trichrome  & Kidney & P & Virtual \\
         \cite{levy2020preliminary} & H\&E & IHC & Liver & P & Virtual \\
         Ours & Dark-field image & H\&E & Breast & P+U & Digital  \\
         
        \bottomrule
    \end{tabular}
}
    \vspace{-4mm}
\label{tab:rela_digital_staining}
\end{table*}

Our main contributions can be summarized as follows:
\begin{itemize}

    \item We collect a novel dataset for digital cell image staining, named DSD, consisting of dark-field unstained and bright-field stained images in the real world with both unpaired and paired-but-misaligned settings.

    \item We propose an effective unpaired digital staining framework reproducible with the datasets and code. A two-stage pipeline to conduct sequentially image enlightening and colorization is introduced. Besides, we utilize the knowledge distillation and cycle-consistency constraint to inject data fidelity and natural image prior.

    \item We further extend the framework to the paired-but-misaligned setting by integrating the LtA module to utilize the pixel-level information between image pairs from adjacent sections.

    \item Proficiently implemented deep digital staining techniques yielding state-of-the-art outcomes for both the unpaired and the paired-but-misaligned settings, surpassing the performance of existing methods.
\end{itemize}    
Our preliminary results have been published as a conference paper\cite{xu2023unsupervised}. The contribution of this journal paper includes:
First, in addition to the original unpaired training data from the DSD dataset in the conference version, we extend our DSD dataset by incorporating the paired-but-misaligned dataset, denoted as DSD\_P.
\Amendment{The detailed dataset settings and collection procedures of the DSD dataset are included.}
Second, we improve the unsupervised framework in the conference version for the unpaired setting by utilizing cycle consistency loss to preserve cells' structural information, which outperforms the previous framework on some metrics and all competing methods.
Third, we introduce a novel paired-but-misaligned training paradigm for the digital staining problem and achieve better performance over competing methods.
Fourth, additional experiments for more competing methods are conducted on the proposed DSD dataset. The results demonstrate the effectiveness of each component in our design.

\section{related work}
\label{sec:related_work}
\subsection{Digital staining}

Digital staining aims to digitally replicate an effect analogous to that achieved through histochemical staining. Many digital staining studies utilize unstained cell images as input, which aligns with our setting and is considered the default configuration for digital staining in our work. 
\Amendmentsecond{Following the definitions in~\cite{rivenson2019phasestain, somani2021digital, mercan2020virtual, lahiani2019virtualization}, we describe the transition from unstained to stained images as digital staining and the transition from one type of staining to another as virtual staining.}
Early studies utilize linear color-coding methods\cite{bautista2011digital,bini2011confocal,bautista2005digital} to predict color, but they suffered from poor generation quality. Learning-based methods have demonstrated superior performance in image processing tasks~\cite{guo2023shadowdiffusion, wang2024progressive, guo2023boundary,ju2024deep,wen2018vidosat}. \Amendmentsecond{As highlighted in recent} \Amendmentsecond{literature reviews~\cite{bai2023deep,kreiss2023digital}, deep learning-based methods have significantly advanced the field of digital staining. } 

A summary of deep learning-based approaches in the field of digital staining is depicted in {\color{cyan}Table~\ref{tab:rela_digital_staining}}. {\color{cyan}Figure~\ref{fig:staining_category}} illustrates the input domain and output domain for various categories of digital staining.

\noindent\textbf{Supervised digital staining.}
\Amendment{Supervised methods assume that their unstained/stained image pairs can be perfectly aligned after some pre-processing techniques, \eg, manually selecting fiducial points and performing the alignment. Previous works have explored various modalities for unstained images, \eg, quantitative phase microscopy images~\cite{rivenson2019phasestain}, autofluorescence images~\cite{zhang2020digital,rivenson2019virtual,rivenson2020emerging,fang2023virtual,doanngan2022label}, and whole slide images (WSI)~\cite{rana2020use}.
Apart from the reconstruction loss, \eg, $\ell_1$ loss, most of them adopted the adversarial loss to achieve better perceptual quality.
However, the pre-processing step for data alignment is a huge burden. 
In this paper, we explore a novel unsupervised framework to relax the data requirements, which involves unpaired and paired-but-misaligned settings.
}

\noindent\textbf{Unsupervised digital staining.}
Recently some unsupervised digital staining methods have been proposed~\cite{xu2023unsupervised,salido2023comparison,kang2022deep,cao2023label}. 
Salido~\etal~\cite{salido2023comparison} compared the performance of CycleGAN~\cite{zhu2017unpaired} and CUT~\cite{park2020contrastive} in digital staining from WSI to H\&E stained images. 
 Cao~\etal~\cite{cao2023label} directly
 utilized CycleGAN, while Kang~\etal~\cite{kang2022deep} incorporated an SSIM loss between the input image and the predicted image using CycleGAN as the baseline.
 These works utilized general methods for unpaired image-to-image translation, overlooking the specifics of digital staining tasks. \Amendment{Based on the CycleGAN baseline, Li~\etal~\cite{UTOM-li2021unsupervised} further introduced a saliency constraint that employs an illumination threshold segmentation method to preserve the background and foreground during translation.}

\noindent\textbf{Virtual staining.} 
Beyond the previously mentioned research, some digital staining techniques leverage a different type of stained image as input to predict the desired type of stained images, commonly referred to as virtual staining~\cite{mercan2020virtual,sun2023bi}, stain transfer~\cite{liu2021unpaired} or digital re-staining~\cite{salido2023comparison}. In this paper, we refer to these tasks as virtual staining.

In clinical practice, pathologists may recommend acquiring additional special stains, following the examination of H\&E stained tissue sections, to improve the accuracy of diagnosis. In situations where certain types of stains have already been prepared and imaged, while other types of stains are needed, virtual staining can significantly reduce the additional labor and costs associated with tissue preparation, staining, and imaging. 
For virtual staining of the unpaired setting, the cycle consistency loss is widely used\cite{mercan2020virtual,lahiani2019virtualization,liu2021unpaired, PEC-lahiani2020seamless, biswas2023generative}. 
Recently some new proposed methods focus on predicting high-resolution stained images \cite{URUST-ho2022ultra, sun2023bi}. \Amendment{Liu~\etal~\cite{liu2021unpaired} employs a segmentation network to assist in training the staining network, but this approach relies on annotations on datasets that may not always be available.} \Amendment{Biswas~\etal~\cite{biswas2023generative} uses hyperspectral H\&E stained images to predict Elastica van Gieson (EVG) stained RGB images.}
Unlike the dark, unstained images in our input domain, stained images demonstrate markedly superior imaging contrast, as shown in {\color{cyan}Figure~\ref{fig:staining_category}}.  Thus contrast-improving methods like low-light image enhancement are not required in their cases. 

\subsection{Low-light enhancement and colorization}  
Digital staining is highly related to low-light image enhancement and colorization.
Numerous low-light image enhancement methods have been proposed, which can be roughly categorized into traditional methods and learning-based approaches. Traditional methods include Histogram Equalization-based methods and Retinex-based methods. Histogram Equalization-based methods\cite{arici2009histogram,lee2013contrast,park2008contrast} extend the dynamic range of a low-light image, while Retinex-based methods\cite{jobson1997properties} decompose a low-light image into reflectance and illuminance maps and consider the reflectance map as the enhanced image. Recent learning-based methods have shown overwhelming results on low-light image enhancement tasks. Wang~\etal~\cite{wang2019underexposed}  introduced an image enhancement network that involves the estimation of an illumination map to connect the low-light input and the anticipated enhancement outcome. Jiang~\etal~\cite{jiang2021enlightengan} proposed an unsupervised method with GAN and attention mechanism.  Zhang~\etal~\cite{zhang2022deep} proposed a color-consistency network aimed at mitigating the color differences between the enhanced image and the ground truth, through a comparison of color histograms of the enhanced images. Ma~\etal~\cite{ma2021structure} proposed an enhancement method for enhancing endoscopic images, corneal confocal microscopy images, and fundus images.

Automatic image colorization seeks to establish a mapping function from a grayscale image to its corresponding colored counterpart, 
which is a classic instance of an ill-posed inverse problem due to the potential existence of multiple color outputs for a single grayscale input.  With the advancements of deep neural networks in recent years, numerous learning-based approaches have been proposed~\cite{zhang2016colorful,zhang2017real,su2020instance,wu2021towards,cao2017unsupervised,vitoria2020chromagan,kumar2021colorization,ji2022colorformer}. For example, earlier endeavors~\cite{zhang2016colorful,zhang2017real} introduced an effective network backbone employing a straightforward pixel-wise $\ell_1$ loss. Subsequently, Wu~\etal~\cite{wu2021towards} incorporated generative color priors as guiding cues. Cao~\etal~\cite{cao2017unsupervised} and Vitoria~\etal~\cite{vitoria2020chromagan} integrated GANs into their approaches to render the results vivid. Su~\etal~\cite{su2020instance} proposed instance-aware colorization, which utilizes previously trained networks for object detection to crop and extract object-level features. Kumar~\etal~\cite{kumar2021colorization} and Ji~\etal~\cite{ji2022colorformer} investigated transformer architectures for this task due to their capacity for non-local modeling.

\begin{figure*}[!t] 
\centering 
\includegraphics[width=1.\textwidth]{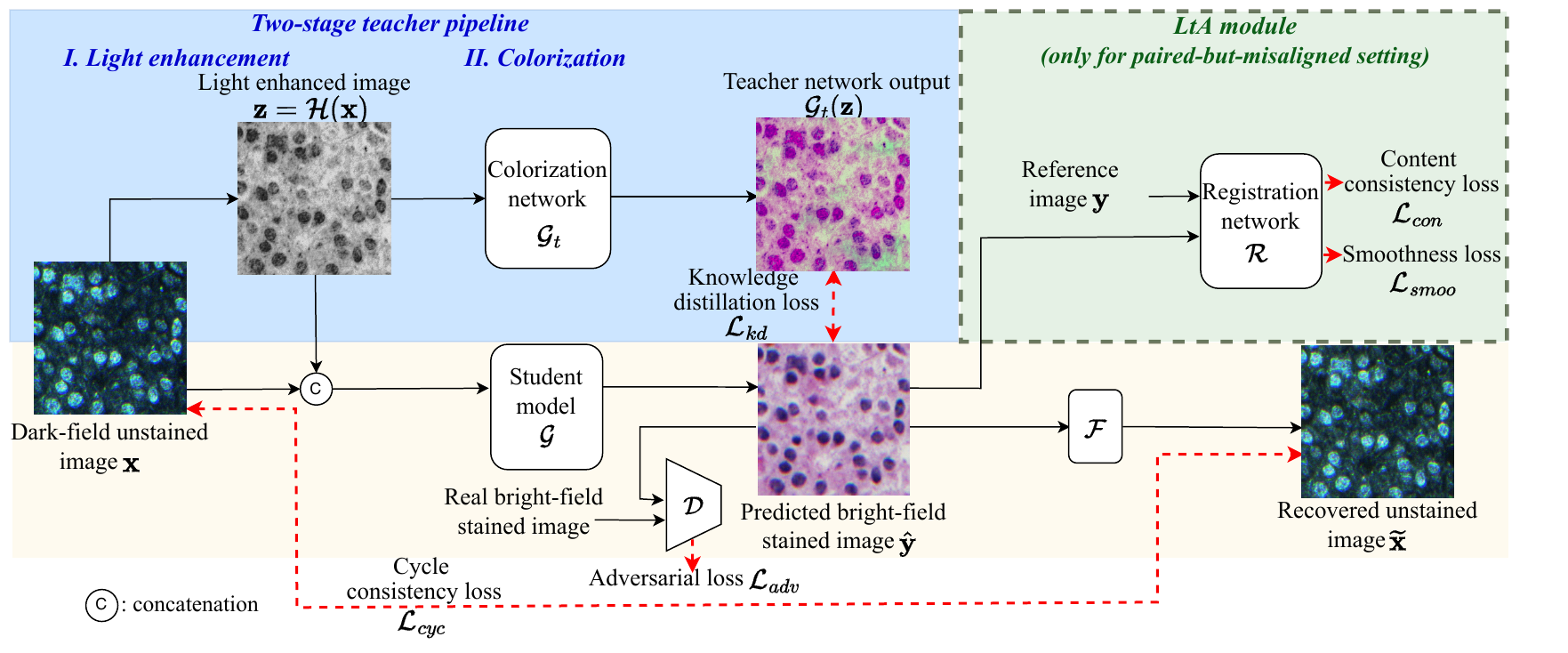}
\vspace{-3mm}
\caption{The overall structure of our method. In our two-stage pipeline, we begin with the light enhancement stage, where the dark-field image $\mathbf{x}$ is processed by $\mathcal{H}$ to obtain the enhanced image $\mathbf{z}$. Then, in the colorization stage, the teacher model produces  $\mathcal{G}_t(\mathbf{z})$ for reference in staining. The student staining generator $\mathcal{G}$ generates the predicted image $\hat{\mathbf{y}}=\mathcal{G}(\mathbf{x} \textcircled{c} \mathbf{z})$.
\Amendmentsecond{The blue, yellow, and green areas indicate Sections IV-A, IV-B, and IV-C, respectively.} }
\vspace{-4mm}
\label{fig:baseline} 
\end{figure*}
\subsection{Knowledge distillation for unpaired I2I translation}
Knowledge distillation aims to enhance the training of a student model through the supervision of a teacher model, which is widely applied in tasks involving medical images~\cite{dou2020unpaired,qin2021efficient}.
Several studies have investigated the application of knowledge distillation in tasks involving unpaired I2I translation~\cite{li2020gan,jin2021teachers,li2020semantic,zhang2022wavelet}. Li~\etal~\cite{li2020gan} minimize the Euclidean distance between the latent features derived from both the teacher and student models.
Jin~\etal~\cite{jin2021teachers} employ distillation on intermediate features of the generator using global kernel alignment. Li~\etal~\cite{li2020semantic} involves generating a semantic relation activation matrix from the feature encoding of the teacher network, which serves as guidance for semantic relation. Zhang~\etal~\cite{zhang2022wavelet} concentrates on distilling the high-frequency bands following discrete wavelet transformation. In our work, knowledge distillation is employed to enhance colorization performance, emphasizing the significance of preserving structural information.

\section{DSD Dataset}
There are some public datasets available for virtual staining such as the ANHIR dataset\cite{borovec2020anhir}. 
However, to the best of our knowledge for digital staining,  especially for dark-field unstained images, there is no publicly available dataset.
So we have collected the DSD dataset, which comprises two subdatasets: DSD\_U and DSD\_P, corresponding to the unpaired and paired-but-misaligned settings, respectively. These images are captured from breast tissue,  all with a field-of-view of 88.4$\mu$m $\times$ 88.4$\mu$m,  corresponding to a resolution of 256$\times$256 pixels.

Our imaging procedures consist of tissue fixation with formalin, paraffin embedding, and sectioning. For DSD\_U, unpaired sections were separated for label-free processing and H\&E staining processing, followed by dark-field imaging and bright-field imaging. 
For DSD\_P, adjacent sections were separated, one was subjected to dark-field imaging, and the other section was subjected to H\&E staining and bright-field imaging to acquire paired images. 
\Amendment{In our experiments, we had to cover specimens with a cover glass for dark-field imaging and bright-field imaging. Removing the cover glass after dark-field imaging proved challenging, so we obtained adjacent vertical sections to capture images with similar structures for the paired-but-misaligned setting.}
\Amendment{We adjusted the light source of the dark-field microscope when collecting DSD\_P samples, resulting in brighter dark-field images in DSD\_P.} Examples from DSD\_U and DSD\_P are shown in {\color{cyan}Figure~\ref{fig:dataset_samples}}.

 For DSD\_U, the training set comprises 559 pairs of unstained dark-field images and stained bright-field images.  These pairs were captured from various tissue structures.
Additionally, to comprehensively assess the performance of our proposed algorithm, we also followed the standard strategy to assemble a testing set consisting of 40 images, matching the resolution of the training images. 
For DSD\_P, our training set comprises 401 image pairs with the same resolution as DSD\_U, while the test dataset comprises 30 image pairs. Moreover, in contrast to the training dataset with misaligned pixels, images in the test dataset are better aligned for a more accurate evaluation. 
\Amendment{We purchased the sample from Precisionmed and used a Thorlabs CS126CU camera for imaging.}
Our dataset can serve as a benchmark for evaluating unsupervised digital staining methods.

\section{Methodology}
In this section, we initially present a two-stage pipeline for the unsupervised digital staining task, which involves histogram matching for light enhancement and image colorization sequentially. By utilizing the pipeline as the teacher model and incorporating the Cycle-GAN structure, we propose our method for the unpaired DSD\_U dataset. For the paired-but-misaligned DSD\_P dataset, we extend our method by adding the LtA module.
The overall structure is illustrated in {\color{cyan}Figure~\ref{fig:baseline}}.

\subsection{Teacher model and assumption}
\label{sec:teacher_model}

To better leverage the prior knowledge of the digital staining task in unsupervised learning, we propose a two-stage staining pipeline for the digital staining task, consisting of the light enhancement stage and image colorization stage.
Specifically, bridging the gap between dark-field and bright-field images involves a two-stage process: initially, enhancing the lightness to improve contrast and generate a grayscale image closely resembling the grayscale representation of the corresponding stained image. Subsequently, colorization is employed to achieve the desired staining style in the image.
\Amendment{
{\color{cyan}Figure~\ref{fig:two_step_examples}} demonstrates the intermediate results in the proposed two-stage framework, where the image contrast and color style progressively are improved at the corresponding stage with consistent structural details such as the positions of cells during the whole process.}
This is a pipeline that guides our approach and enables us to propose a teacher model predicting a stained image as a reference.
\subsubsection{Light enhancement}
\label{sec:light_enhancement}

\begin{figure}[!t] 
\centering 
\includegraphics[width=0.48\textwidth]{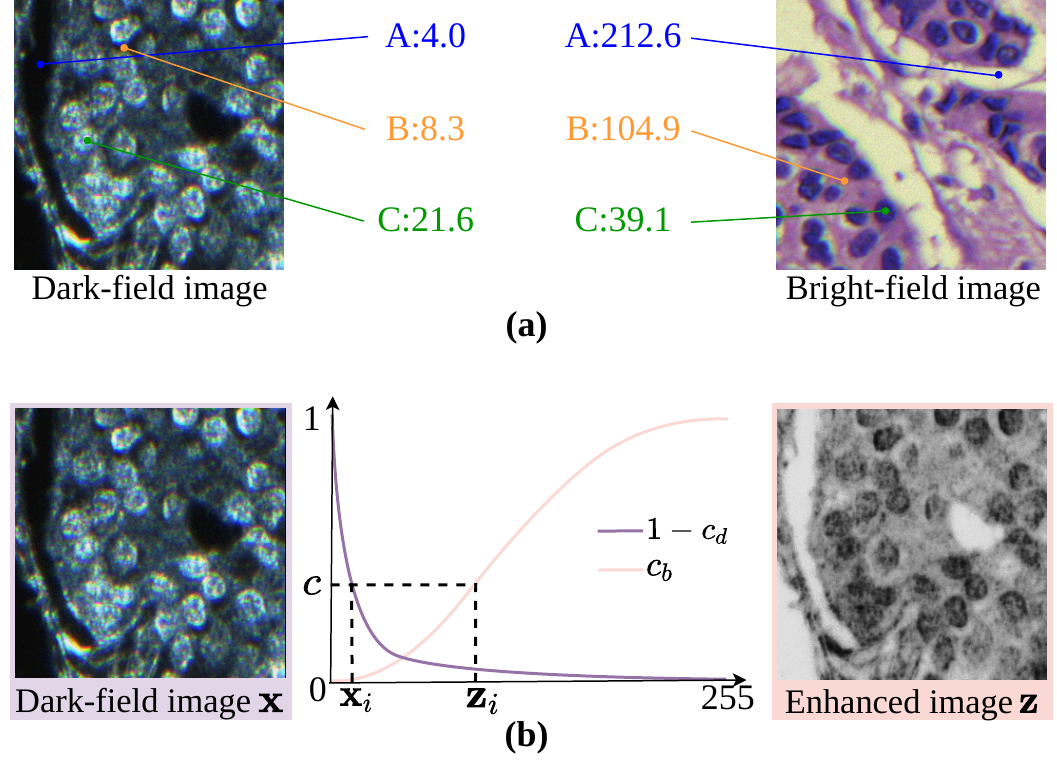} 
\caption{(a). Two example images from dark-field images and bright-field images of DSD\_U. A, B, and C denote pairs of pixels located in the cavity area, pixels next to cells, and pixels inside cells respectively. A, B, and C's illuminance intensity values are shown in the figure as the values range from 0 to 255. (b). Histogram matching from dark-field to enhanced images, with $c=1-c_d(\mathbf{x}_i)=c_b(\mathbf{z}_i)$. }
\vspace{-5mm}
\label{fig:his_matching} 
\end{figure}

In dark-field unstained images, cell regions typically appear brighter than the background, while in bright-field stained images, they appear darker than the background, as shown in {\color{cyan}Figure~\ref{fig:his_matching}(a)}. 
This observation leads us to the assumption that the illuminance relationship between cell regions in dark-field unstained images and bright-field stained images is approximately reversed. 
To address this, we propose matching the illuminance distribution of enhanced outputs to that of bright-field images.

Histogram matching is a well-established technique widely used in image processing tasks, such as camera color correction\cite{ding2020multi} and SDR-to-HDR imaging and tone mapping~\cite{bihan2019inverse}. In our task, we apply this concept by inversely mapping the illuminance distribution of dark-field images to that of bright-field images, in accordance with the previously mentioned assumption.
\Amendment{We denote $c_b$ and $c_d$ as the cumulative distribution functions of the illuminance distribution for bright-field image $\mathbf{y}$ and dark-field image $\mathbf{x}$, respectively.}
For a pixel illuminance value $\mathbf{x}_i$ from $\mathbf{x}$, where $i$ represents pixel coordinates, let $\mathbf{z}_i$ denote the grayscale intensity of the corresponding pixel in the enhanced image $\mathbf{z}$. We establish the relationship as $c_b(\mathbf{z}_i)=1-c_d(\mathbf{x}_i)$. Consequently, we compute $\mathbf{z}_i$ utilizing the inverse cumulative distribution function as $\mathbf{z}_i=c_b^{-1}(1-c_d(\mathbf{x}_i))$.
This intensity mapping enables us to produce enhanced cell images denoted as $\mathbf{z}=\mathcal{H}(\mathbf{x})$, leading to significantly enhanced visibility compared to dark-field unstained images, as illustrated in {\color{cyan}Figure~\ref{fig:his_matching}(b)}.

\subsubsection{Colorization}

In the CIE $LAB$ color space, a RGB image with shape $H\times W$ can be decomposed into the luminance $\mathbf{y}_{l}\in \mathbb{R}^{1\times H \times W}$ and the chrominance $\mathbf{y}_{ab}\in \mathbb{R}^{2\times H \times W}$. Grayscale image colorization is the process of estimating the absent chrominance $\mathbf{y}_{ab}$ from the provided luminance $\mathbf{y}_{l}$ in order to reconstruct the full RGB image.
We generate synthetic gray-color training pairs by inversely transforming existing H\&E bright-field stained images into their gray counterparts. These synthetic image pairs serve as training data for training the colorization network $\mathcal{G}_t$, using a pixel-wise $\ell_1$ loss. 

\Amendment{
In summary, we designed an unsupervised teacher model for digital staining. This model uses a non-learning histogram matching method for light enhancement and trains a colorization network using synthesized gray images. Although these steps are optimized individually and not globally, the teacher model serves as a foundation for developing an end-to-end}
\Amendment{student model that can further learn structural information.
} 
\subsection{Student model for unsupervised staining}
\label{sec:distillation}

\begin{figure}[!t] 
\vspace{-5mm}
\centering
\includegraphics[width=0.5\textwidth]{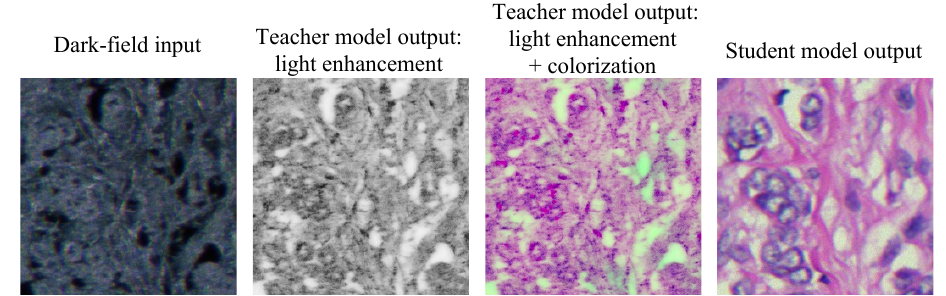}
\caption{Illustration of the light enhancement and colorization stages for an example from DSD\_P.} 
\vspace{-5mm}
\label{fig:two_step_examples} 
\end{figure}

For the straightforward two-stage pipeline, it has a limited effect as shown in {\color{cyan}Figure~\ref{fig:two_step_examples}}. 
To mitigate the gap between the stained images $\mathcal{G}_t(\mathbf{z})$ predicted by the two-stage pipeline and real-stained images, what we proposed is to utilize the two-stage pipeline as the teacher model and train an end-to-end staining model using distillation. 

 The trained teacher model predicts $\mathbf{y}_t$ from dark-field image $\mathbf{x}$ through light enhancement and colorization, which accurately captures the cells' position, shape, and desired staining style. Simultaneously, the staining generator $\mathcal{G}$ takes the unstained image $\mathbf{x}$ concatenated with $\mathbf{z}$ as input to predict the digitally stained image $\hat{\mathbf{y}}$.

To ensure that the knowledge learned from the teacher model is effectively transferred to the generator, we employ a knowledge distillation loss denoted as $\mathcal{L}_{kd}$. This loss is designed to minimize the $\ell_1$ distance between $\mathbf{y}_t$ and $\hat{\mathbf{y}}$, and its definition is as follows:
\begin{equation}
\begin{aligned}
    \mathcal{L}_{kd} & = \|\hat{\mathbf{y}}-\mathbf{y}_t\|_1 \\
    &= \|\mathcal{G}(\mathbf{x} \textcircled{c} \mathcal{H}(\mathbf{x}))-\mathcal{G}_t(\mathbf{z})\|_1 \;.
\end{aligned}
\end{equation}
Note that, we solely employ the distilled student staining network for inference.

The cycle consistency loss, which was originally proposed to regularize mapping functions in image-to-image translation tasks~\cite{zhu2017unpaired} has found widespread application in digital staining tasks~\cite{salido2023comparison,kang2022deep,cao2023label}.
In our task, we incorporate this loss to further constrain structural information.
 We introduce a generator $\mathcal{F}$ that maps from bright-field stained modality $\mathcal{Y}$ to dark-field unstained modality $\mathcal{X}$. 
For a dark-field unstained image $\mathbf{x}$ subjected to forward-backward translations, self-consistency is maintained: $\mathcal{F}(\hat{\mathbf{y}})=\mathcal{F}(\mathcal{G}(\mathbf{x}\textcircled{c} \mathcal{H}(\mathbf{x}))) \approx \mathbf{x}$ for $\mathbf{x} \in \mathcal{X}$.  This adherence to self-consistency leads to the formulation of the corresponding cycle consistency loss $\mathcal{L}_{cyc}$:
\begin{equation}
\mathcal{L}_{cyc}=\|\mathcal{F}(\hat{\mathbf{y}})-\mathbf{x}\|_1 . 
\end{equation} 

Simultaneously, we apply a least-square GAN (LSGAN)~\cite{mao2017least}  which replaces the binary cross-entropy terms of the original GAN objectives with least squares loss functions, following our previous work~\cite{xu2023unsupervised} and other works on medical images like~\cite{hu2021bidirectional,colleoni2022ssis}. In this setup, we introduce a discriminator $\mathcal{D}$ to discriminate between samples drawn from digitally stained images $\hat{\mathbf{y}}$ and true stained images $\mathbf{y}$. The discriminator $\mathcal{D}$ is trained to classify histochemically stained samples $\mathbf{y}$ as 1 and samples synthesized from the generator $\hat{\mathbf{y}}$ as 0. Meanwhile, the generator $\mathcal{G}$ is trained to deceive the discriminator.
This approach encourages $\hat{\mathbf{y}}$ to closely resemble histochemically stained images $\mathbf{y}$ and follow the same distribution as $\mathbf{y}$.
The adversarial loss functions, denoted as $\mathcal{L}_{adv}$ for the generator $\mathcal{G}$ and $\mathcal{L}_D$ for the discriminator $\mathcal{D}$, are as follows:

\begin{equation}
\begin{aligned}
    \mathcal{L}_{adv} &= (\mathcal{D}(\hat{\mathbf{y}})-1)^2 \\
    \mathcal{L}_{D} &=  (\mathcal{D}(\mathbf{y})-1)^2+\mathcal{D}^2(\hat{\mathbf{y}})
\end{aligned}
\end{equation}

In the unpaired setting, the total loss function for staining generator $\mathcal{G}$ is calculated as follows:
\begin{equation}
    \mathcal{L}_U= \mathcal{L}_{adv}+\lambda_1\mathcal{L}_{kd}+\lambda_2\mathcal{L}_{cyc} ,
\end{equation}
where $\lambda_1$ and $\lambda_2$ represent weights for each loss item.

\subsection{Extension to the paired-but-misaligned setting}
\label{sec:registration_part} 

\begin{figure}[!t] 
\centering
\vspace{-5mm}
\includegraphics[width=0.5\textwidth]{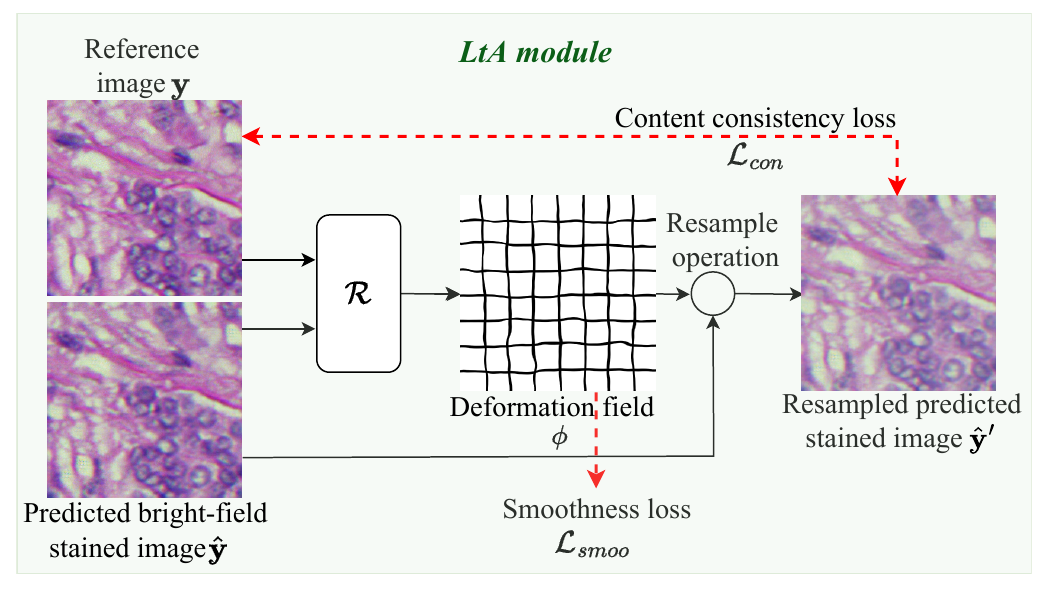}
\caption{The structure for the LtA module. In the paired-but-misaligned setting, where the stained reference image $\mathbf{y}$ is available, albeit misaligned, we employ a registration network $\mathcal{R}$ to generate a deformation field $\phi$ to align $\mathbf{y}$ and $\hat{\mathbf{y}}$, resulting in a resampled image $\hat{\mathbf{y}}'$.} 
\vspace{-4mm}
\label{fig:registration_part} 
\end{figure}

During the data preparation stage, we discovered a practical method for generating paired dark-field and bright-field images. By using adjacent sections, with one for dark-field imaging and the other for staining and bright-field imaging, we easily obtain a pair of dark-field unstained images and bright-field stained images. While these two images are somewhat paired, there is no guarantee of pixel alignment, as illustrated in {\color{cyan}Figure~\ref{fig:dataset_samples}}.
We further effectively utilize pixel-wise information from these misaligned image pairs by incorporating the Learning to Align (LtA) module inspired by RegGAN~\cite{kong2021breaking}, into the existing components designed for the unpaired setting. The LtA module registrates images from domain $\mathcal{Y}$ by modeling the geometric distortion between images from adjacent sections.

\begin{figure*}[!t] 
\centering 
\includegraphics[width=1.\textwidth]{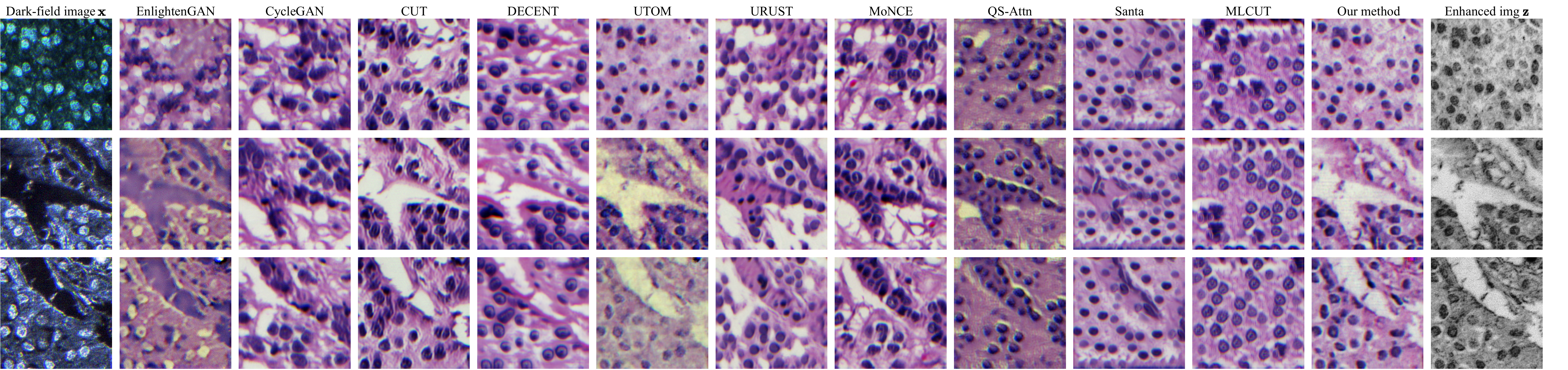} 
\caption{
Visual examples of digital staining results from the unpaired dataset DSD\_U using our proposed method and competing methods.  } 
\vspace{-6mm}
\label{fig:journal_unpaired_competing} 
\end{figure*}

First the registration network $\mathcal{R}$ generates a deformation field $\phi=R(\hat{\mathbf{y}}, \mathbf{y})$ that is required to align the predicted image $\hat{\mathbf{y}}=\mathcal{G}((\mathbf{x} \textcircled{c} \mathcal{H}(\mathbf{x})))$ with the reference image $\mathbf{y}$. Then we apply a resampling operation to $\hat{\mathbf{y}}$ using the deformation field $\phi$ to generate a registered stained image $\hat{\mathbf{y}}'=\hat{\mathbf{y}} \circ \phi $, where $\circ$ denotes the resampling operation.  The deformation field $\phi$ is in the form of $\mathbb{R}^{2\times H \times W}$, where each pixel in an image of size $H\times W$ has two channels, indicating the horizontal and vertical displacement, respectively. 

Since image style and geometric structure both contribute to the overall visual appearance, learning geometric changes becomes challenging when image styles undergo significant changes. To mitigate this challenge, we first perform image translation to handle style changes.  Then, we feed images from the stained domain into the LtA module, enabling $\mathcal{R}$ to concentrate on learning geometric distortions effectively, while the staining generator $\mathcal{G}$ focuses on capturing the staining style. The corresponding content consistency loss $\mathcal{L}_{con}$ is defined as follows:
\begin{equation}
    \mathcal{L}_{con} = \|\hat{\mathbf{y}}'-\mathbf{y}\|_1.
\end{equation}
For the deformation field, we adopt a smoothness loss $\mathcal{L}_{smoo}$ utilizing the isotropic squared Total Variation (TV) loss\cite{aly2005image} of $\phi$ to minimize the gradient of $\phi$:
\begin{equation}
    \mathcal{L}_{smoo} = \frac{1}{WH}\sum_{i,j} (||(\nabla_{h}\phi)_{ij}||_2+||(\nabla_{v}\phi)_{ij}||_2) ,
\end{equation}
where $\nabla_{h}$ and $\nabla_{v}$ denote the horizontal and vertical gradient operators, respectively. $W$ and $H$ represent the width and height of the image, and $i,j$ indicates the pixel coordinates, $1\leq i \leq H, 1\leq j \leq W$.

The total loss function in the paired-but-misaligned setting equals to:
\begin{equation}
    \mathcal{L}_P= \mathcal{L}_U+\lambda_3\mathcal{L}_{con}+\lambda_4\mathcal{L}_{smoo} ,
\end{equation}
where $\lambda_3$ and $\lambda_4$ represent weights for each loss item added for the paired-but-misaligned setting.

\section{Experiments}
\noindent \textbf{Implementation details.} 
The generator network $\mathcal{G}$ is constructed using basic blocks from ResNet. $\mathcal{G}$ includes one convolutional layer, two 2-stride convolutional layers, nine residual blocks, and two fractionally strided convolutions with a stride of $\frac{1}{2}$, followed by one convolutional layer, inspired by~\cite{zhu2017unpaired}, which has proven effective in various image-to-image translation tasks. In our work, we set the number of input channels to 4 and the number of output channels to 3.  The network $\mathcal{F}$ shares the same structure as $\mathcal{G}$, except that the number of input and output channels is both set to 3. For the colorization network $\mathcal{G}_t$, we employ a U-Net-based architecture following~\cite{su2020instance}. $\mathcal{G}_t$ is trained separately for the two settings using H\&E stained images from the training datasets in DSD\_U and DSD\_P. \Amendment{As the structure of the discriminator network $\mathcal{D}$, following CycleGAN~\cite{zhu2017unpaired}, Pix2Pix~\cite{isola2017image} and RegGAN~\cite{kong2021breaking}, we use a network with five layers of convolutions for the discriminator.  The structure of the registration network $\mathcal{R}$ follows RegGAN.
All images are linearly rescaled to the range [-1, 1].
} 
Regarding the value of hyper-parameters, in the unpaired setting, we set $\lambda_1=5$ and $\lambda_2=80$; in the paired-but-misaligned setting, we use $\lambda_1=2$, $\lambda_2=10$,  $\lambda_3=20$ and $\lambda_4=10$.
For optimization, we utilize the Adam optimizer for loss minimization for 300 epochs, with an initial learning rate of 0.0001 for the first 200 epochs. Subsequently, we perform linear decay of the learning rate to 0 over the last epochs.

\noindent \textbf{Evaluation metrics.} 
\Amendmentsecond{We aim for the predicted images to exhibit staining styles similar to real stained images while preserving the structural information of the input unstained images.
In the unpaired setting, for the staining style, we employ a set of perceptual metrics for evaluation including Frechet Inception Distance (FID)~\cite{heusel2017gans}, Kernel Inception Distance (KID)~\cite{bińkowski2018demystifying}, and Natural Image Quality Evaluator (NIQE)~\cite{mittal2012making} for evaluation. These metrics are widely used metrics in unpaired image-to-image translation~\cite{park2020contrastive, xie2022unsupervised, xie2023unpaired} and virtual staining~\cite{sun2023bi, URUST-ho2022ultra}}.
\Amendmentsecond{As we have a limited number of real stained images in the test set, which may lack sufficient diversity for these metrics, we calculate these metrics between real stained bright-field images from the training set in DSD\_U and predicted stained images from the testing set. }
\Amendmentsecond{While these metrics evaluate distribution similarity between predicted and reference images, they do not assess structural similarity, which is critical for our task.}
Additionally, we propose a new metric, using a variation of Learned Perceptual Image Patch Similarity (LPIPS)~\cite{zhang2018unreasonable}, denoted as LPIPS-c in our work, to assess the model's ability to preserve content information from the input image. Specifically, we calculate LPIPS-c as the LPIPS value between the grayscale image of the predicted bright-field image $\hat{\mathbf{y}}_{gray}$ and the corresponding enhanced image $\mathbf{z}$. In all these metrics, a lower value indicates better visual quality.
We also add computational consumption (FLOPs) and the number of parameters as metrics to compare the computational efficiency between models.

In the unpaired setting, to further evaluate model performance, we conducted a new user preference study, as automatic metrics like FID and KID do not fully align with the requirements of digital staining. Specifically, we carried out a comparison-based user study in which participants were presented with structural features from both unstained and stained images, along with real stained image examples to illustrate the target staining style. Participants were asked to select the best-matched image from the options — one that closely resembled the real stained images in style while preserving the structural information of the input unstained images. If no clear preference was identified, participants were allowed to select up to two matching images. 
For each comparison between competing methods and in the ablation study, we randomly selected 20 unstained images from the testing set. 
\Amendmentthird{For each unstained image, stained images were generated using all methods, including ours, and participants were asked to select the best one, forming a single question with the order of options randomized.
Responses were collected from 20 participants with sufficient medical background knowledge. After excluding one incomplete response, the user preference rates were summarized based on the 19 valid responses. }

\Amendment{In the paired-but-misaligned setting, Peak Signal to Noise Ratio (PSNR),  Structural Similarity (SSIM), and LPIPS were used as metrics to evaluate the performances of trained models based on the testing dataset of DSD\_P.}
All these metrics are calculated between predicted images and reference images. Higher PSNR and SSIM values indicate better performance, while a lower LPIPS value indicates better performance. \Amendment{We used the AlexNet network pretrained on the BAPPS dataset to calculate the LPIPS and LPIPS-c metrics, following the default settings.}
\subsection{Digital staining over DSD\_U dataset}

\begin{table}[t]
\centering
\footnotesize
\setlength{\tabcolsep}{0.2em}
\renewcommand{\arraystretch}{0.9}
\caption{Quantitative evaluation of the digital staining results on the DSD$\_$U dataset. The results achieving the best and second-best rankings under each metric are highlighted in {\color{red}\textbf{Red}} and {\color{blue}Blue}, respectively. \Amendmentthird{380 denotes 19 participants $\times$ 20 questions}.}
\adjustbox{width=1.0\linewidth}{
        \begin{tabular}{l|c|ccc|cc|c}
        \toprule
          Method & LPIPS-c$\downarrow$ & NIQE$\downarrow$ &  FID$\downarrow$ &KID$\downarrow$ & \makecell[c]{FLOPs \\(G)$\downarrow$} & \makecell[c]{\# Param \\ (M)$\downarrow$} & \makecell[c]{User Preference\\ Rate$\uparrow$}\\\midrule
         \rowcolor{Gray}Ours & {\color{red}\textbf{0.339}}&{\color{red}\textbf{4.960}} & {\color{red}\textbf{147.6}} & {\color{red}\textbf{0.087}} & 114.01 & 11.38 &{\color{red}\textbf{46.05\% (175/380)}} \\\midrule
         UTOM~\cite{UTOM-li2021unsupervised} &{\color{blue}0.428} &{\color{blue}8.526}&218.3&0.189 & 36.28 & 54.41 & {\color{blue}18.29\% (69.5/380)}\\
         URUST~\cite{URUST-ho2022ultra}& 0.662&10.017&158.9&0.125 & 206.48 & 13.43& 1.58\% (6/380)\\\midrule
        EnlightenGAN~\cite{jiang2021enlightengan} &0.469& 24.707 &172.5&  0.099& 32.88 & 8.64& 3.16\% (12/380)\\
         CycleGAN~\cite{zhu2017unpaired} &  0.654  & 8.962 & 150.0 & {\color{blue}0.094} & 113.60 & 11.38 & 0.39\% (1.5/380)\\ 
                  CUT~\cite{park2020contrastive} &0.595 &20.746& 164.7  & 0.097& 128.56 & 11.38  & 15.53\% (59/380)\\ 
                  DECENT~\cite{xie2022unsupervised} &0.679&17.406&158.7&0.120& 128.56 & 11.38& 12.89\% (49/380)\\
                  Santa~\cite{xie2023unpaired} & 0.625 & 14.604 & 215.03 & 0.224 & 59.27 & 11.43 & 0.26\% (1/380)\\
                  MLCUT~\cite{han2023multilevel} & 0.591 & 14.501 & 205.44 & 0.192 & 128.56 & 11.38 &0.26\% (1/380)\\
                  MoNCE~\cite{zhan2022modulated} & 0.700 & 17.891 & {\color{blue}149.2} & 0.092& 128.56 & 11.38& 1.32\% (5/380)\\
                  QS-Attn~\cite{hu2022qs} & 0.641& 15.557 &189.6 & 0.176 & 128.56 & 11.38& 0.26\% (1/380)\\
        \bottomrule
    \end{tabular}
}
    
    \vspace{-4mm}
\label{tab:compe_unpair_table}
\end{table}

For unpaired digital staining methods~\cite{salido2023comparison,kang2022deep,cao2023label}, even though they have not released their code, they have claimed to use CUT~\cite{park2020contrastive} and CycleGAN~\cite{zhu2017unpaired}. 
As for unpaired virtual and digital staining methods, we have considered URUST~\cite{URUST-ho2022ultra} and UTOM~\cite{UTOM-li2021unsupervised} as competing methods. 
\Amendmentsecond{Additionally, we have selected several recent unpaired image-to-image (I2I) translation methods, namely EnlightenGAN~\cite{jiang2021enlightengan}, CycleGAN~\cite{zhu2017unpaired}, CUT~\cite{park2020contrastive},  DECENT~\cite{xie2022unsupervised}, MoNCE~\cite{zhan2022modulated}, QS-Attn~\cite{hu2022qs}, and Santa~\cite{xie2023unpaired} as competing methods.} To ensure a fair and equitable comparison, we re-trained all competing methods on the training set of our DSD\_U dataset.
{\color{cyan}Table~\ref{tab:compe_unpair_table}} shows the quantitative experimental results, in which the proposed method outperforms all competing methods for image quality.

\begin{table}[!t]
\centering
\footnotesize
\setlength{\tabcolsep}{0.2em}
\renewcommand{\arraystretch}{0.9}
\caption{Quantitative comparison of the digital staining results on DSD\_P. }
\adjustbox{width=0.9\linewidth}{
        \begin{tabular}{l|ccc|cc}
        \toprule
         Method & PSNR$\uparrow$ & SSIM$\uparrow$ &  LPIPS$\downarrow$  & FLOPs (G)$\downarrow$ & \# Param (M)$\downarrow$\\\midrule
         \rowcolor{Gray}Ours & {\color{red}\textbf{19.939}}&{\color{red}\textbf{0.374}} &{\color{red}\textbf{0.273}} & 114.01 & 11.38\\\midrule
         Pix2Pix~\cite{isola2017image} &18.833 &0.297&0.347&36.29&54.41\\
         RegGAN~\cite{kong2021breaking}& 19.238&0.346& {\color{blue}0.292}&113.60&11.38\\
        SwinIR~\cite{liang2021swinir} &{\color{blue}19.659}& 0.337 &0.311&145.69&0.91\\
         Palette~\cite{saharia2022palette} &  17.362  & 0.369 & 0.383&409.66&62.64 \\ 

        HER2-virtual~\cite{doanngan2022label} &19.244 &0.353& 0.366&253.98&36.46\\ 
                 DINO~\cite{DINO} & 19.277 & {\color{blue}0.372} & 0.401 &21.06 &37.50 \\
        \bottomrule
    \end{tabular}
}
    
    \vspace{-4mm}
\label{tab:compe_misaligned_table}
\end{table}

\begin{figure*}[t] 
\vspace{-5mm}
\centering 
\includegraphics[width=1\textwidth]{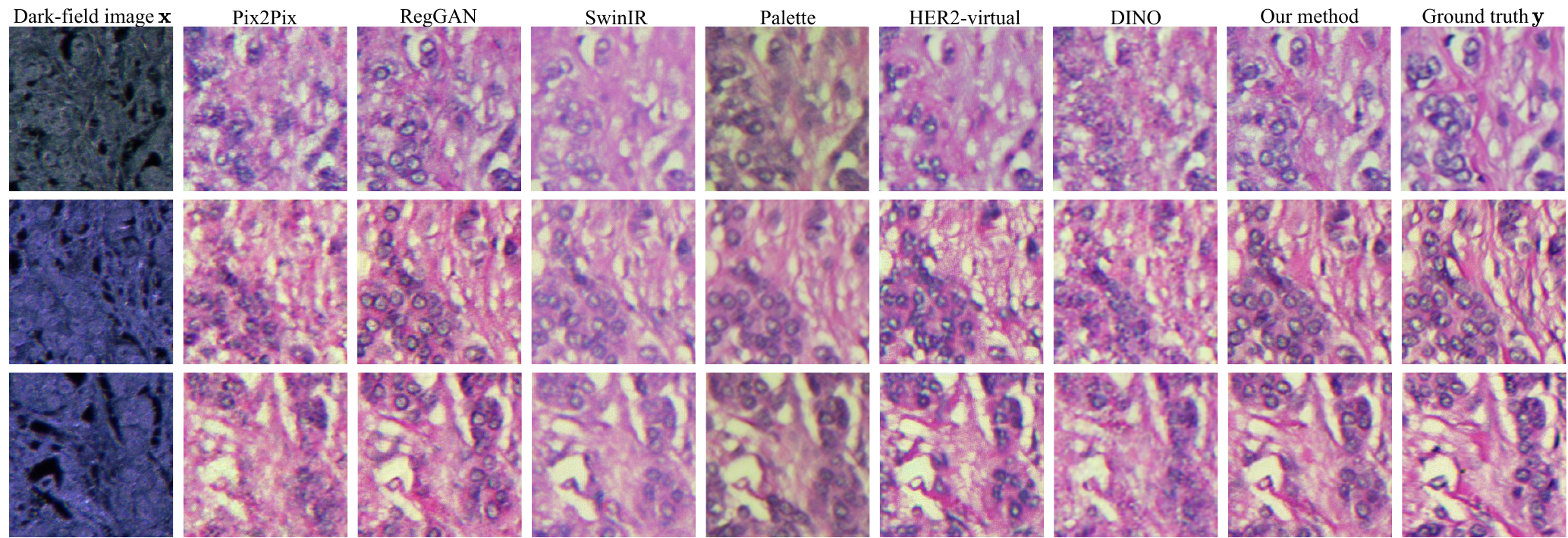} 
\caption{
Examples of digital staining results from DSD\_P using our proposed method and other competing methods.}
\vspace{-4mm}
\label{fig:journal_paired_competing} 
\end{figure*}
A visualization of the unpaired staining results of our proposed method and competing methods on the DSD\_U dataset is shown in {\color{cyan}Figure~\ref{fig:journal_unpaired_competing}}. EnlightenGAN, CycleGAN, and URUST fail to distinguish between cells and background regions, leading to erroneous predictions of cavities instead of cells. CUT produces cells with rough edges and artificial cavity areas. In the second example, UTOM predicts a darker and more blurred stained image, while in the third example, it predicts a lighter color for the cell region compared to the output image from our method and normal H\&E stained images. Since DECENT focuses on encouraging density to be consistent for all patches in the translation, it successfully distinguishes cells from cavities. However, its results still demonstrate issues such as indistinct cell edges and artificial cavity areas. \Amendmentsecond{Santa fails to predict the cavity area, while MLCUT predicts incomplete, hollow cells.}
\Amendment{MoNCE and QS-Attn, both based on contrastive learning, also fail to distinguish between cell and background regions.}
In contrast, our method consistently generates stained bright-field images with complete structural information, including the accurate position and shape of cells and cavities, while maintaining a high-quality staining style. 
\Amendment{Our model's parameters and computational requirements are comparable to other methods and do not exhibit significant differences, while our method demonstrates better performance.}

\subsection{Digital staining over DSD\_P dataset}

To the best of our knowledge, no code for digital staining work with the paired setting has been released, and we have only identified one virtual staining method with available code, referred to as HER2-virtual~\cite{doanngan2022label}, which we incorporated as a competing method.  We also considered paired I2I translation methods such as Pix2Pix~\cite{isola2017image}, Palette~\cite{saharia2022palette}, and DINO~\cite{DINO}, image restoration methods like SwinIR~\cite{liang2021swinir} for comparison.
It is worth noting that Palette utilizes a diffusion network, while SwinIR leverages a ViT network. 
Additionally, we employed RegGAN~\cite{kong2021breaking}, an image-to-image translation method featuring a LtA module to deal with misaligned datasets, as another competing approach.
We retrained these competing methods using the training set of our dataset DSD\_P. 
{\color{cyan}Table~\ref{tab:compe_misaligned_table}} clearly demonstrates the quantitative results, showing that our proposed method surpasses all of the competing methods in terms of both pixel fidelity and perceptual fidelity. Furthermore, we have provided visual examples in  {\color{cyan}Figure~\ref{fig:journal_paired_competing}}.  Specifically, SwinIR~\cite{liang2021swinir} exhibited an abnormal staining style in its predictions. Pix2Pix produced images with indistinct cell regions. \Amendment{Palette~\cite{saharia2022palette}, HER2-virtual~\cite{doanngan2022label}, RegGAN~\cite{kong2021breaking}, and DINO~\cite{DINO} all predicted cell positions and shapes less accurately compared to our method}. 
\Amendment{While our model's parameters and computational requirements are comparable to other methods and show no significant differences, our method achieves better performance.}

\subsection{Ablation study}
\noindent \textbf{Effects of $\mathcal{L}_{kd}$ and $\mathcal{L}_{cyc}$.} 

\begin{figure}[!t] 
\centering 
\includegraphics[width=0.48\textwidth]{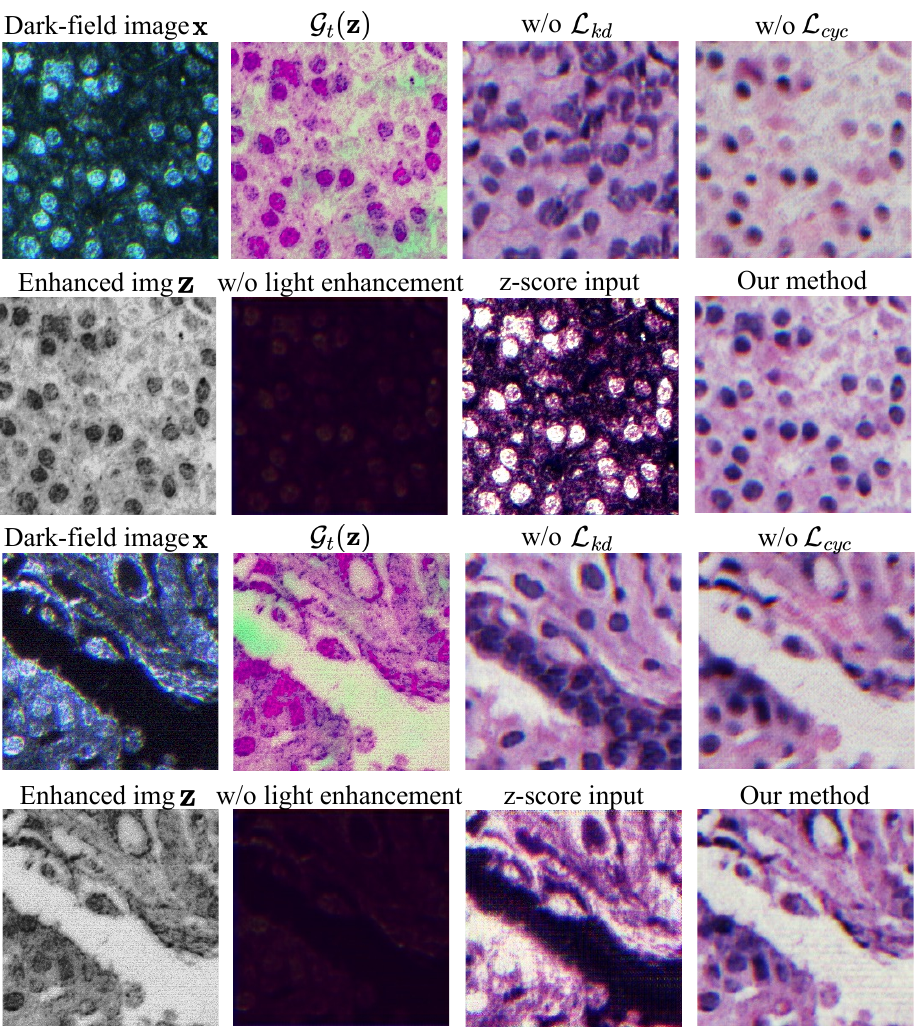} 
\caption{Visual comparison of ablation study for digital staining results over examples from unpaired dataset DSD\_U.} 
\vspace{-4mm}
\label{fig:journal_unpaired_ablation} 
\end{figure}

We thoroughly investigate the impact of each loss function item applied in the training stage. For the unpaired dataset DSD\_U, {\color{cyan}Table~\ref{tab:ablation_unpair_table}} presents the quantitative evaluation results and {\color{cyan}Figure~\ref{fig:journal_unpaired_ablation}} demonstrates the visual examples with different loss functions. 

First, we verify the effectiveness of the knowledge distillation loss $\mathcal{L}_{kd}$, 
in which we remove the $\mathcal{L}_{kd}$ and preserve the cycle consistency loss $\mathcal{L}_{cyc}$ and adversarial loss $\mathcal{L}_{adv}$. 
According to the quantitative results shown in {\color{cyan}Table~\ref{tab:ablation_unpair_table}}, there was an improvement in FID and KID metrics, but NIQE and LPIPS metrics suffered a drop without $\mathcal{L}_{kd}$. 
\Amendment{FID and KID are metrics comparing the similarity of image distributions between two sets, regardless of the preservation of structural information.}
However, from {\color{cyan}Figure~\ref{fig:journal_unpaired_ablation}} we noticed that it predicts the cell part and background part reversely like the CycleGAN. 
\Amendment{This is coherent with the conclusion from Cohen~\etal~\cite{cohen2018distribution}, which states that using only cycle-consistency loss and adversarial loss for unsupervised medical image translation can cause image features to hallucinate, potentially adding or removing structural details.}
This highlights the necessity of $\mathcal{L}_{kd}$ to help the model distinguish between cell and background parts and to predict with the correct staining style.

Then we investigate the influence of the cycle consistency loss $\mathcal{L}_{cyc}$. According to {\color{cyan}Table~\ref{tab:ablation_unpair_table}}, the model's performance is worse than the proposed methods in terms of all quantitative metrics. Qualitatively, the visual examples from {\color{cyan}Figure~\ref{fig:journal_unpaired_ablation}} revealed that removing $\mathcal{L}_{cyc}$ resulted in predicted stained images with oversmoothed edges for cells. Additionally, these images tended to predict a lighter color for some cells, similar to the predicted output of the teacher model but different from normal H\&E stained images. This suggests that $\mathcal{L}_{cyc}$ plays an important role in preserving more structural details for the position and shapes of cells.

\begin{figure}[t] 
\centering 
\includegraphics[width=0.5\textwidth]{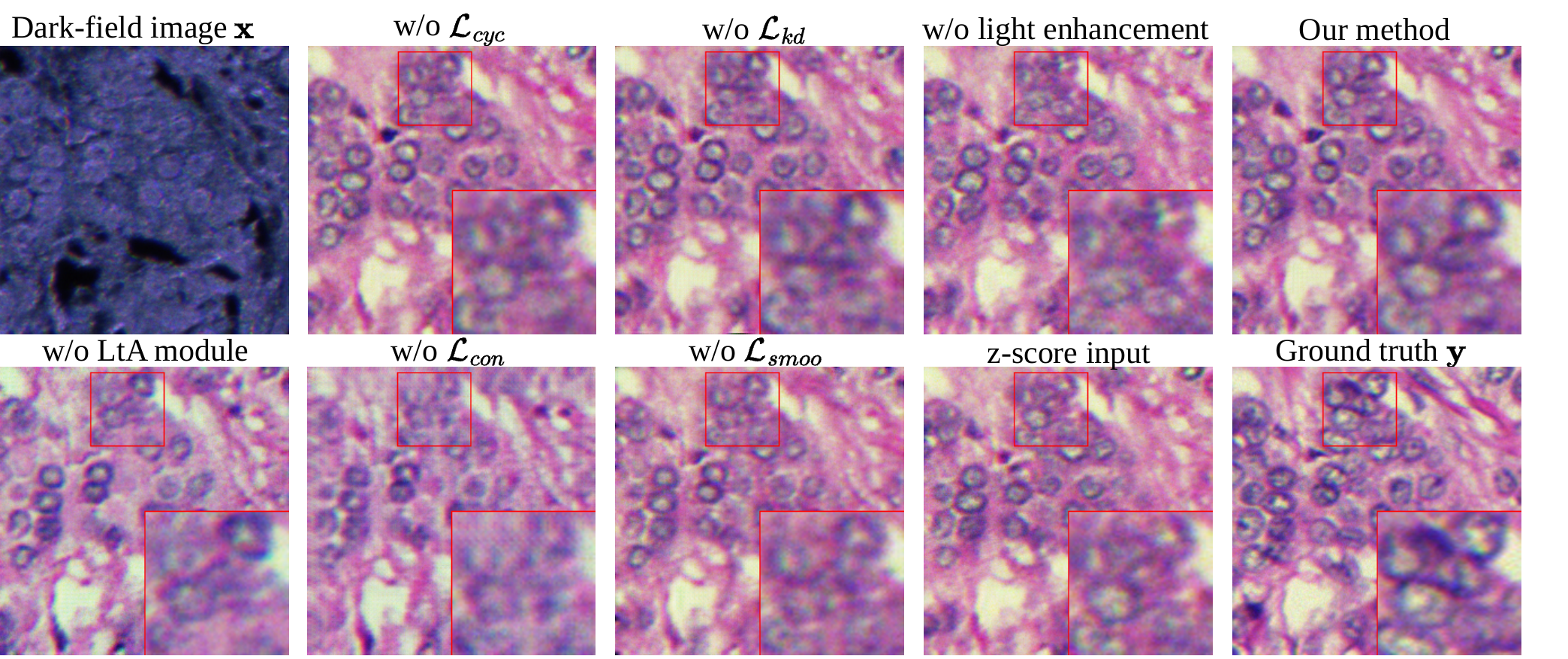} 
\caption{Visual comparison of ablation study for digital staining results
over examples from unpaired dataset DSD\_P.} 
\label{fig:dsd_p_ablation_img} 
\end{figure}

\Amendment{For the paired-but-misaligned setting, we also conducted experiments where we excluded the knowledge distillation loss $\mathcal{L}_{kd}$ and the cycle consistency loss $\mathcal{L}_{cyc}$ separately, as shown in {\color{cyan}Figure~\ref{fig:dsd_p_ablation_img}} and {\color{cyan}Table~\ref{tab:ablation_misaligned_table}}. These experiments demonstrated a noticeable deterioration in performance across all the evaluation metrics.  }

\begin{table}[!t]
\centering
\footnotesize
\renewcommand{\arraystretch}{0.8}
\caption{Quantitative evaluation of the digital staining results for ablation study on DSD\_P. }
\adjustbox{width=.8\linewidth}{
    \begin{tabular}{l|cccc}
        \toprule
         Method & PSNR$\uparrow$ & SSIM$\uparrow$ &  LPIPS$\downarrow$ \\\midrule
         \rowcolor{Gray}Ours & {\color{red}\textbf{19.939}}&{\color{red}\textbf{0.374}} &{\color{red}\textbf{0.273}}  \\\midrule
         w/o LtA module &17.647&0.302&0.353\\
         w/o $\mathcal{L}_{con} $ & 17.755 & 0.317 & 0.346\\
         w/o $\mathcal{L}_{smoo} $ & 19.306 & 0.346 & 0.302\\
         w/o $\mathcal{L}_{kd}$ &19.481&0.339&{\color{blue}0.287}\\
         w/o $\mathcal{L}_{cyc}$& {\color{blue}19.779}&{\color{blue}0.358}&0.291\\
         \midrule
         w/o light enhancement & 19.544 &0.351 &0.295\\
         z-score input & 19.550 & 0.356 & 0.293 \\
        \bottomrule
    \end{tabular}
}
    
\vspace{-4mm}
\label{tab:ablation_misaligned_table}
\end{table}
\begin{table}[!t]
\centering
\footnotesize
\renewcommand{\arraystretch}{0.8}
\caption{Quantitative evaluation of the digital staining results for ablation study on DSD\_U. }
\adjustbox{width=1.0\linewidth}{
    \begin{tabular}{l|c|ccc|c}
        \toprule
         Method & LPIPS-c$\downarrow$ & NIQE$\downarrow$ &  FID$\downarrow$ &KID$\downarrow$ & \makecell[c]{User Preference\\ Rate$\uparrow$}\\\midrule
        \rowcolor{Gray}Ours & {\color{red}\textbf{0.339}}&{\color{red}\textbf{4.960}} &{\color{blue}147.6} & {\color{blue}0.087} &{\color{red}\textbf{70.39\% (267.5/380)}} \\\midrule
         w/o $\mathcal{L}_{kd}$ &0.460 &{\color{blue}5.045}&{\color{red}\textbf{137.5}}&{\color{red}\textbf{0.068}} & 7.50\% (28.5/380)\\
         w/o $\mathcal{L}_{cyc}$& {\color{blue}0.403}&10.00&164.7&0.105& {\color{blue}17.37\% (66/380)}\\
         \midrule
         w/o light enhancement &0.923 &26.052&367.5&0.415& 0.00\% (0/380)\\
         z-score input &0.502 &12.570&182.2&0.091& 4.74\% (18/380)\\
        \bottomrule
    \end{tabular}
}
    
    \vspace{-5mm}
\label{tab:ablation_unpair_table}
\end{table}

\noindent \textbf{Effects of the LtA module for the paired-but-misaligned setting.} 
For the paired-but-misaligned dataset DSD\_P, to certify our motivation for incorporating the LtA module, we conducted a comparative analysis between our method and a variant in which we removed the LtA module. This variant essentially follows the same approach as our proposed method for the unpaired dataset. 
\Amendment{The quantitative results of this comparison are summarized in {\color{cyan}Table~\ref{tab:ablation_misaligned_table}} and the visual examples are demonstrated in {\color{cyan}Figure~\ref{fig:dsd_p_ablation_img}} with different loss functions.}
\Amendment{For the LtA module added in the paired-but-misaligned setting, removing the content consistency loss $\mathcal{L}_{con}$ or the entire LtA module resulted in a PSNR drop of approximately 2dB, leading to blurry and distorted cells, highlighting the importance of pixel-wise supervision. Performance also declined in the absence of the smoothness loss $\mathcal{L}_{smoo}$. }
\Amendment{Our method predicts colored cell images with the clearest and most accurate structural information while preserving the finest staining style among all settings, thereby confirming the effectiveness of these components.}

\begin{table*}[!t]
\centering
\footnotesize
\renewcommand{\arraystretch}{1.1}
\caption{\Amendment{Quantitative comparison of our methods across various settings using our synthetic White Blood Cell (WBC) dataset. All the classification metrics are computed using a ConvNeXt\_Tiny classification network pre-trained on the original WBC dataset.} }
\adjustbox{width=0.7\linewidth}{
    \begin{tabular}{l|cc|cccc}
        \toprule
        Input   & PSNR$\uparrow$ & SSIM$\uparrow$ &    
           Acc.$\uparrow$ &  F-m. $\uparrow$ & Pre. $\uparrow$ & Rec.$\uparrow$ \\\midrule
           Synthesized unstained images & - & - & 0.751 & 0.615 & 0.723 & 0.641 \\
           \midrule
          Stained by Ours (Paired-but-misaligned)  & 36.21 & 0.955 &0.939 & 0.902 & 0.902 & 0.904\\
         Stained by Ours (Unpaired)   & 33.20 & 0.941 &0.930 & 0.889 & 0.890 & 0.892\\
        \midrule
      Physical stained images  & - & - & 0.949 & 0.922 & 0.923 & 0.923 \\
        \bottomrule
    \end{tabular}
}
    
    \vspace{-4mm}
\label{tab:wbc_performance_metric}
\end{table*}

\noindent \textbf{Effects of light enhancement step.} 
\Amendment{We further explore the effect of the proposed light enhancement step in our method. The purpose of the light enhancement step is to produce the light enhanced image $\mathbf{z}$, where the illuminance is reversed as described in \textbf{Section IV-A.1}, which is necessary for the following colorization step. First, we remove the light enhancement step and use dark-field images as the input for the colorization network $\mathcal{G}_t$ and the student model $\mathcal{G}$.} 
\Amendment{The quantitative comparisons of removing light enhancement step are shown in   {\color{cyan}Table~\ref{tab:ablation_unpair_table}} and {\color{cyan}Table~\ref{tab:ablation_misaligned_table}}. 
From {\color{cyan}Figure~\ref{fig:journal_unpaired_ablation}} we noticed that this setting predicts excessively dark images in the unpaired setting. 
The performance dropped dramatically after removing the light enhancement step. 
}

\Amendment{Second, since one of the effects of the light enhancement step is contrast enhancement, we conducted the ablation study to replace the light enhancement step with z-score normalization, using dark-field images after z-score normalization as the input for the colorization network $\mathcal{G}_t$ and the student model $\mathcal{G}$.}
\Amendment{The performance of z-score normalization is worse than our proposed method, in both the unpaired and the paired-but-misaligned settings, as shown in {\color{cyan}Table~\ref{tab:ablation_unpair_table}} and {\color{cyan}Table~\ref{tab:ablation_misaligned_table}}. As shown in {\color{cyan}Figure~\ref{fig:journal_unpaired_ablation}}, this setting also predicts the cell
part and background part reversely in the unpaired setting. In conclusion, the light enhancement step plays a major role in our method. We did not observe significant improvement in the competing methods when using the light enhanced image $\mathbf{z}$ as input. Since they are not co-designed with the light enhancement step, light enhancement is an ad-hoc module for them.
}

\subsection{Downstream diagnostic task using digital staining}

\begin{figure}[!t] 
\centering
\includegraphics[width=0.5\textwidth]{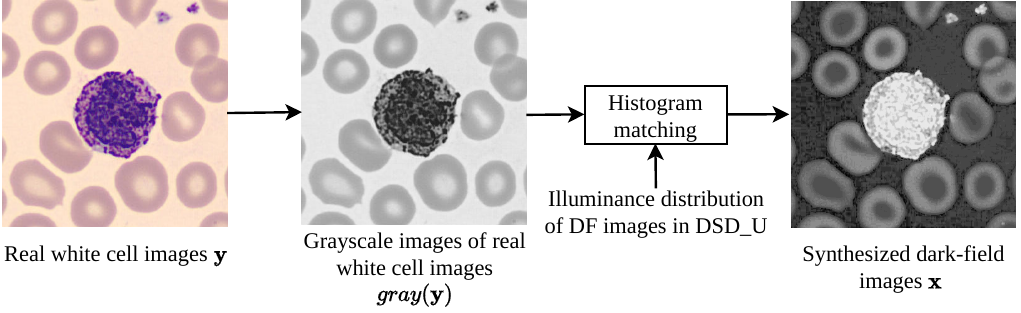}
\caption{Illustration of synthesis of dark-field images of WBC images.} 
\label{fig:wbc_systhesize} 
\end{figure}

\Amendment{
We tried to extend our digital staining methods to other cell images with annotations to further examine our digital staining methods in a real clinic. However, to the best of our knowledge, there are no public datasets with dark-field images and corresponding annotations. We compromise to use synthesized data, choose a dataset with labels, and synthesize dark-field images from given bright-field images. Given the White Blood Cell (WBC) dataset with morphological attributes}
\Amendment{from \cite{tsutsui2024wbcatt}, with each cell annotated with 11 attributes and its corresponding cell type (monocyte, lymphocyte, neutrophil, eosinophil, basophil).  
We try to synthesize corresponding dark-field images $\mathbf{x}$ using histogram matching, following our assumption proposed in {\color{cyan}Section~\ref{sec:light_enhancement}}, as shown in {\color{cyan}Figure~\ref{fig:wbc_systhesize}}. We calculate grayscale images of real white cell } \Amendment{images $gray(\mathbf{y})$ from real white cell images $\mathbf{y}$. Following the assumption that the illuminance relationship between cell regions in dark-field unstained images and bright-field stained images is reversed, }
\Amendment{we use $\mathbf{x}_i=c_d^{-1}(1-c_b(gray(\mathbf{y})_i))$ to get synthesized dark-field images $\mathbf{x}$,  where $i$ represents pixel coordinates, and $c_b$ and $c_d$ denotes the cumulative distribution}
\Amendment{functions of the illuminance distribution for real stained WBC images $\mathbf{y}$ and dark-field image $\mathbf{x}$ from DSD\_U respectively.}

\Amendment{
We train staining models from $\mathbf{x}$ to white cell images $\mathbf{y}$ using our proposed methods for both the paired setting and the}
\Amendment{unpaired setting. We obtain the synthesized enhanced images$\mathbf{z}$ from synthesized dark-field images $\mathbf{x}$ and the illuminance distribution of $gray(\mathbf{y})$, following the light enhancement step we stated in {\color{cyan}Section~\ref{sec:light_enhancement}}. 
}

\Amendment{
 Apart from image accuracy metrics like PSNR and SSIM, we also use classification metrics such as Macro Accuracy (Acc.), Macro F-measure (F-m.), Macro Precision (Pre.), and Macro Recall (Rec.), where the metrics are unweighted mean for every label from all 11 attributes and its cell type. All the classification metrics are computed using a ConvNeXt\_Tiny classification network pre-trained on the original WBC dataset.  From {\color{cyan}Table~\ref{tab:wbc_performance_metric}}, the accuracy for predicted stained images only dropped for 1\% to 2\% for both unpaired and paired settings, which is valuable for clinical diagnosis.}

\subsection{Inference time on large images}
 \Amendmentsecond{

 Since the inference process relies entirely on the student model, a fully convolutional network, it can theoretically scale to any size, including whole-slide images. However, hardware memory constraints impose practical limits on the maximum image size for a single inference run. 
{\color{cyan}Table~\ref{tab:size_gpu}} provides a detailed analysis of inference efficiency, showing the time required and memory usage for input image sizes ranging from 256 to 5000 pixels on specific GPUs. The average inference time does not scale up dramatically. This demonstrates the computational scalability and resource demands of our digital staining process.
For even larger images, we can follow the approach described in~\cite{sun2023bi}, performing inference on overlapping patches and combining the results. Using this method, it is possible to process a $50,000 \times 50,000$ image in approximately 800 seconds on an RTX 3080 machine.
}
 \begin{table}[!t]
\centering
\footnotesize
\renewcommand{\arraystretch}{0.8}
\caption{Inference time and GPU memory usage across different GPU models for large image sizes on DSD\_U.}
\adjustbox{width=1.0\linewidth}{
    \begin{tabular}{l|c|c|c|c|c}
        \toprule
         Image size (pixels per side) & $256$ & $1000$ &  $2000$ & $4000$ & $5000$\\\midrule
         Nvidia GPU model & RTX3080 & RTX3080 & RTX3080 & A5000 & A100 \\\midrule
         Average inference time (s) & 0.157 & 0.411 & 1.381 & 4.473 & 2.724\\\midrule
         GPU memory usage (GB)& 1.10 & 3.63 & 9.16 & 23.09 & 30.39\\
        \bottomrule
    \end{tabular}
}
    
\vspace{-4mm}
\label{tab:size_gpu}
\end{table}

\section{Conclusion}

In this paper, we introduce the Digital Staining Dataset (DSD), which comprises two subdatasets: DSD\_U for unpaired data and DSD\_P for paired-but-misaligned data. We present an innovative framework for unsupervised digital staining employing knowledge distillation, catering to both unpaired and paired-but-misaligned settings. In contrast to previous techniques that rely on direct image-to-image translation methods in unpaired settings, our approach excels at accurately predicting structural attributes, such as cell positions and shapes, while achieving the desired staining style. After obtaining image pairs with paired-but-misaligned stained and unstained images from adjacent sections, we extend our methodology to address the paired-but-misaligned setting by introducing the LtA module. Leveraging pixel-level information between misaligned image pairs, this addition significantly enhances the performance of our approach for DSD\_P when compared to the proposed method for the unpaired setting. Our extensive experimental results on DSD illustrate the superior quantitative and qualitative outcomes achieved by our proposed methods in comparison to competing methods. 
\Amendment{We extended our digital staining method to the WBC (White Blood Cell) dataset, exploring potential medical applications.}

\bibliographystyle{IEEEtran}
\bibliography{refs}

\begin{thebibliography}{10}
\providecommand{\url}[1]{#1}
\csname url@samestyle\endcsname
\providecommand{\newblock}{\relax}
\providecommand{\bibinfo}[2]{#2}
\providecommand{\BIBentrySTDinterwordspacing}{\spaceskip=0pt\relax}
\providecommand{\BIBentryALTinterwordstretchfactor}{4}
\providecommand{\BIBentryALTinterwordspacing}{\spaceskip=\fontdimen2\font plus
\BIBentryALTinterwordstretchfactor\fontdimen3\font minus \fontdimen4\font\relax}
\providecommand{\BIBforeignlanguage}[2]{{%
\expandafter\ifx\csname l@#1\endcsname\relax
\typeout{** WARNING: IEEEtran.bst: No hyphenation pattern has been}%
\typeout{** loaded for the language `#1'. Using the pattern for}%
\typeout{** the default language instead.}%
\else
\language=\csname l@#1\endcsname
\fi
#2}}
\providecommand{\BIBdecl}{\relax}
\BIBdecl

\bibitem{feldman2014tissue}
A.~T. Feldman and D.~Wolfe, ``Tissue processing and hematoxylin and eosin staining,'' in \emph{Histopathology}.\hskip 1em plus 0.5em minus 0.4em\relax Springer, 2014, pp. 31--43.

\bibitem{rivenson2019phasestain}
Y.~Rivenson, T.~Liu, Z.~Wei, Y.~Zhang, K.~de~Haan, and A.~Ozcan, ``Phasestain: the digital staining of label-free quantitative phase microscopy images using deep learning,'' \emph{Light: Science \& Applications}, vol.~8, no.~1, pp. 1--11, 2019.

\bibitem{cheng2023disentangled}
H.~Cheng, Y.~Wang, H.~Li, A.~C. Kot, and B.~Wen, ``Disentangled feature representation for few-shot image classification,'' \emph{IEEE transactions on neural networks and learning systems}, vol.~35, no.~8, pp. 10\,422--10\,435, 2024.

\bibitem{zhang2020digital}
Y.~Zhang, K.~de~Haan, Y.~Rivenson, J.~Li, A.~Delis, and A.~Ozcan, ``Digital synthesis of histological stains using micro-structured and multiplexed virtual staining of label-free tissue,'' \emph{Light: Science \& Applications}, vol.~9, no.~1, pp. 1--13, 2020.

\bibitem{rivenson2019virtual}
Y.~Rivenson, H.~Wang, Z.~Wei, K.~de~Haan, Y.~Zhang, Y.~Wu, H.~G{\"u}nayd{\i}n, J.~E. Zuckerman, T.~Chong, A.~E. Sisk \emph{et~al.}, ``Virtual histological staining of unlabelled tissue-autofluorescence images via deep learning,'' \emph{Nature biomedical engineering}, vol.~3, no.~6, pp. 466--477, 2019.

\bibitem{rivenson2020emerging}
Y.~Rivenson, K.~de~Haan, W.~D. Wallace, and A.~Ozcan, ``Emerging advances to transform histopathology using virtual staining,'' \emph{BME Frontiers}, vol. 2020, 2020.

\bibitem{fang2023virtual}
Z.~Fang, R.~Kozikowski, K.~de~Haan, S.~Alexanian, M.~E. Kallen, A.~Rosenbloom, C.~Glaser, M.~Conner, Y.~Liang, K.~Teplitz \emph{et~al.}, ``Virtual staining overlay enabled combined morphological and spatial transcriptomic analysis of individual malignant b cells and local tumor microenvironments,'' in \emph{Medical Imaging with Deep Learning, short paper track}, 2023.

\bibitem{doanngan2022label}
B.~DoanNgan, D.~Angus, L.~Sung \emph{et~al.}, ``Label-free virtual her2 immunohistochemical staining of breast tissue using deep learning,'' \emph{BME frontiers}, 2022.

\bibitem{rana2020use}
A.~Rana, A.~Lowe, M.~Lithgow, K.~Horback, T.~Janovitz, A.~Da~Silva, H.~Tsai, V.~Shanmugam, A.~Bayat, and P.~Shah, ``Use of deep learning to develop and analyze computational hematoxylin and eosin staining of prostate core biopsy images for tumor diagnosis,'' \emph{JAMA network open}, vol.~3, no.~5, pp. e205\,111--e205\,111, 2020.

\bibitem{kang2022deep}
L.~Kang, X.~Li, Y.~Zhang, and T.~T. Wong, ``Deep learning enables ultraviolet photoacoustic microscopy based histological imaging with near real-time virtual staining,'' \emph{Photoacoustics}, vol.~25, p. 100308, 2022.

\bibitem{cao2023label}
R.~Cao, S.~D. Nelson, S.~Davis, Y.~Liang, Y.~Luo, Y.~Zhang, B.~Crawford, and L.~V. Wang, ``Label-free intraoperative histology of bone tissue via deep-learning-assisted ultraviolet photoacoustic microscopy,'' \emph{Nature Biomedical Engineering}, vol.~7, no.~2, pp. 124--134, 2023.

\bibitem{tomczak2023digital}
A.~Tomczak, S.~Ilic, G.~Marquardt, T.~Engel, N.~Navab, and S.~Albarqouni, ``Digital staining of white blood cells with confidence estimation,'' \emph{IEEE Transactions on Medical Imaging}, 2023.

\bibitem{verebes2013hyperspectral}
G.~S. Verebes, M.~Melchiorre, A.~Garcia-Leis, C.~Ferreri, C.~Marzetti, and A.~Torreggiani, ``Hyperspectral enhanced dark field microscopy for imaging blood cells,'' \emph{Journal of biophotonics}, vol.~6, no. 11-12, pp. 960--967, 2013.

\bibitem{chua2022label}
J.~J. Chua, J.~C.~Y. Xin, S.~Zhang, and M.~Olivo, ``Label-free pathology imaging with multimodal hyperspectral microscopy for breast cancer diagnosis,'' in \emph{Clinical and Translational Biophotonics}.\hskip 1em plus 0.5em minus 0.4em\relax Optica Publishing Group, 2022, pp. JM3A--49.

\bibitem{zhang2023label}
S.~Zhang, S.~Zeng, W.~Liao, R.~R.~Z. Tan, and M.~Olivo, ``Label-free cancer classification using hyperspectral imaging microscopy and machine learning,'' in \emph{The European Conference on Lasers and Electro-Optics}.\hskip 1em plus 0.5em minus 0.4em\relax Optica Publishing Group, 2023, p. cl\_p\_4.

\bibitem{xu2023unsupervised}
Z.~Xu, L.~Guo, S.~Zhang, A.~C. Kot, and B.~Wen, ``Unsupervised deep digital staining for microscopic cell images via knowledge distillation,'' in \emph{ICASSP 2023-2023 IEEE International Conference on Acoustics, Speech and Signal Processing (ICASSP)}.\hskip 1em plus 0.5em minus 0.4em\relax IEEE, 2023, pp. 1--5.

\bibitem{liu2021unpaired}
S.~Liu, B.~Zhang, Y.~Liu, A.~Han, H.~Shi, T.~Guan, and Y.~He, ``Unpaired stain transfer using pathology-consistent constrained generative adversarial networks,'' \emph{IEEE Transactions on Medical Imaging}, vol.~40, no.~8, pp. 1977--1989, 2021.

\bibitem{salido2023comparison}
J.~Salido, N.~Vallez, L.~Gonz{\'a}lez-L{\'o}pez, O.~Deniz, and G.~Bueno, ``Comparison of deep learning models for digital h\&e staining from unpaired label-free multispectral microscopy images,'' \emph{Computer Methods and Programs in Biomedicine}, vol. 235, p. 107528, 2023.

\bibitem{UTOM-li2021unsupervised}
X.~Li, G.~Zhang, H.~Qiao, F.~Bao, Y.~Deng, J.~Wu, Y.~He, J.~Yun, X.~Lin, H.~Xie \emph{et~al.}, ``Unsupervised content-preserving transformation for optical microscopy,'' \emph{Light: Science \& Applications}, vol.~10, no.~1, p.~44, 2021.

\bibitem{pillar2022virtual}
N.~Pillar and A.~Ozcan, ``Virtual tissue staining in pathology using machine learning,'' \emph{Expert Review of Molecular Diagnostics}, vol.~22, no.~11, pp. 987--989, 2022.

\bibitem{pradhan2021computational}
P.~Pradhan, T.~Meyer, M.~Vieth, A.~Stallmach, M.~Waldner, M.~Schmitt, J.~Popp, and T.~Bocklitz, ``Computational tissue staining of non-linear multimodal imaging using supervised and unsupervised deep learning,'' \emph{Biomedical Optics Express}, vol.~12, no.~4, pp. 2280--2298, 2021.

\bibitem{lahiani2019virtualization}
A.~Lahiani, J.~Gildenblat, I.~Klaman, S.~Albarqouni, N.~Navab, and E.~Klaiman, ``Virtualization of tissue staining in digital pathology using an unsupervised deep learning approach,'' in \emph{Digital Pathology: 15th European Congress, ECDP 2019, Warwick, UK, April 10--13, 2019, Proceedings 15}.\hskip 1em plus 0.5em minus 0.4em\relax Springer, 2019, pp. 47--55.

\bibitem{PEC-lahiani2020seamless}
A.~Lahiani, I.~Klaman, N.~Navab, S.~Albarqouni, and E.~Klaiman, ``Seamless virtual whole slide image synthesis and validation using perceptual embedding consistency,'' \emph{IEEE Journal of Biomedical and Health Informatics}, vol.~25, no.~2, pp. 403--411, 2020.

\bibitem{sun2023bi}
K.~Sun, Z.~Chen, G.~Wang, J.~Liu, X.~Ye, and Y.-G. Jiang, ``Bi-directional feature fusion generative adversarial network for ultra-high resolution pathological image virtual re-staining,'' in \emph{Proceedings of the IEEE/CVF Conference on Computer Vision and Pattern Recognition}, 2023, pp. 3904--3913.

\bibitem{URUST-ho2022ultra}
M.-Y. Ho, M.-S. Wu, and C.-M. Wu, ``Ultra-high-resolution unpaired stain transformation via kernelized instance normalization,'' in \emph{European Conference on Computer Vision}.\hskip 1em plus 0.5em minus 0.4em\relax Springer, 2022, pp. 490--505.

\bibitem{biswas2023generative}
T.~Biswas, H.~Suzuki, M.~Ishikawa, N.~Kobayashi, and T.~Obi, ``Generative adversarial network based digital stain conversion for generating rgb evg stained image from hyperspectral h\&e stained image,'' \emph{Journal of Biomedical Optics}, vol.~28, no.~5, pp. 056\,501--056\,501, 2023.

\bibitem{de2021deep}
K.~de~Haan, Y.~Zhang, J.~E. Zuckerman, T.~Liu, A.~E. Sisk, M.~F. Diaz, K.-Y. Jen, A.~Nobori, S.~Liou, S.~Zhang \emph{et~al.}, ``Deep learning-based transformation of h\&e stained tissues into special stains,'' \emph{Nature communications}, vol.~12, no.~1, p. 4884, 2021.

\bibitem{levy2020preliminary}
J.~J. Levy, C.~R. Jackson, A.~Sriharan, B.~C. Christensen, and L.~J. Vaickus, ``Preliminary evaluation of the utility of deep generative histopathology image translation at a mid-sized nci cancer center,'' \emph{bioRxiv}, pp. 2020--01, 2020.

\bibitem{somani2021digital}
A.~Somani, A.~A. Sekh, I.~S. Opstad, {\AA}.~B. Birgisdottir, T.~Myrmel, B.~S. Ahluwalia, K.~Agarwal, D.~K. Prasad, and A.~Horsch, ``Digital staining of mitochondria in label-free live-cell microscopy,'' in \emph{Bildverarbeitung f{\"u}r die Medizin 2021: Proceedings, German Workshop on Medical Image Computing, Regensburg, March 7-9, 2021}.\hskip 1em plus 0.5em minus 0.4em\relax Springer, 2021, pp. 235--240.

\bibitem{mercan2020virtual}
C.~Mercan, G.~Mooij, D.~Tellez, J.~Lotz, N.~Weiss, M.~van Gerven, and F.~Ciompi, ``Virtual staining for mitosis detection in breast histopathology,'' in \emph{2020 IEEE 17th International Symposium on Biomedical Imaging (ISBI)}.\hskip 1em plus 0.5em minus 0.4em\relax IEEE, 2020, pp. 1770--1774.

\bibitem{bautista2011digital}
P.~A. Bautista and Y.~Yagi, ``Digital staining for histopathology multispectral images by the combined application of spectral enhancement and spectral transformation,'' in \emph{2011 Annual International Conference of the IEEE Engineering in Medicine and Biology Society}.\hskip 1em plus 0.5em minus 0.4em\relax IEEE, 2011, pp. 8013--8016.

\bibitem{bini2011confocal}
J.~Bini, J.~Spain, K.~Nehal, V.~Hazelwood, C.~DiMarzio, and M.~Rajadhyaksha, ``Confocal mosaicing microscopy of human skin ex vivo: spectral analysis for digital staining to simulate histology-like appearance,'' \emph{Journal of biomedical optics}, vol.~16, no.~7, pp. 076\,008--076\,008, 2011.

\bibitem{bautista2005digital}
P.~A. Bautista, T.~Abe, M.~Yamaguchi, Y.~Yagi, and N.~Ohyama, ``Digital staining of unstained pathological tissue samples through spectral transmittance classification,'' \emph{Optical review}, vol.~12, pp. 7--14, 2005.

\bibitem{guo2023shadowdiffusion}
L.~Guo, C.~Wang, W.~Yang, S.~Huang, Y.~Wang, H.~Pfister, and B.~Wen, ``Shadowdiffusion: When degradation prior meets diffusion model for shadow removal,'' in \emph{Proceedings of the IEEE/CVF Conference on Computer Vision and Pattern Recognition}, 2023, pp. 14\,049--14\,058.

\bibitem{wang2024progressive}
C.~Wang, L.~Guo, Y.~Wang, H.~Cheng, Y.~Yu, and B.~Wen, ``Progressive divide-and-conquer via subsampling decomposition for accelerated mri,'' in \emph{Proceedings of the IEEE/CVF Conference on Computer Vision and Pattern Recognition}, 2024, pp. 25\,128--25\,137.

\bibitem{guo2023boundary}
L.~Guo, C.~Wang, W.~Yang, Y.~Wang, and B.~Wen, ``Boundary-aware divide and conquer: A diffusion-based solution for unsupervised shadow removal,'' in \emph{Proceedings of the IEEE/CVF International Conference on Computer Vision}, 2023, pp. 13\,045--13\,054.

\bibitem{ju2024deep}
Y.~Ju, K.-M. Lam, W.~Xie, H.~Zhou, J.~Dong, and B.~Shi, ``Deep learning methods for calibrated photometric stereo and beyond,'' \emph{IEEE Transactions on Pattern Analysis and Machine Intelligence}, 2024.

\bibitem{wen2018vidosat}
B.~Wen, S.~Ravishankar, and Y.~Bresler, ``Vidosat: High-dimensional sparsifying transform learning for online video denoising,'' \emph{IEEE Transactions on Image processing}, vol.~28, no.~4, pp. 1691--1704, 2018.

\bibitem{bai2023deep}
B.~Bai, X.~Yang, Y.~Li, Y.~Zhang, N.~Pillar, and A.~Ozcan, ``Deep learning-enabled virtual histological staining of biological samples,'' \emph{Light: Science \& Applications}, vol.~12, no.~1, p.~57, 2023.

\bibitem{kreiss2023digital}
L.~Kreiss, S.~Jiang, X.~Li, S.~Xu, K.~C. Zhou, K.~C. Lee, A.~M{\"u}hlberg, K.~Kim, A.~Chaware, M.~Ando \emph{et~al.}, ``Digital staining in optical microscopy using deep learning-a review,'' \emph{PhotoniX}, vol.~4, no.~1, p.~34, 2023.

\bibitem{zhu2017unpaired}
J.-Y. Zhu, T.~Park, P.~Isola, and A.~A. Efros, ``Unpaired image-to-image translation using cycle-consistent adversarial networks,'' in \emph{Proceedings of the IEEE international conference on computer vision}, 2017, pp. 2223--2232.

\bibitem{park2020contrastive}
T.~Park, A.~A. Efros, R.~Zhang, and J.-Y. Zhu, ``Contrastive learning for unpaired image-to-image translation,'' in \emph{European conference on computer vision}.\hskip 1em plus 0.5em minus 0.4em\relax Springer, 2020, pp. 319--345.

\bibitem{arici2009histogram}
T.~Arici, S.~Dikbas, and Y.~Altunbasak, ``A histogram modification framework and its application for image contrast enhancement,'' \emph{IEEE Transactions on image processing}, vol.~18, no.~9, pp. 1921--1935, 2009.

\bibitem{lee2013contrast}
C.~Lee, C.~Lee, and C.-S. Kim, ``Contrast enhancement based on layered difference representation of 2d histograms,'' \emph{IEEE transactions on image processing}, vol.~22, no.~12, pp. 5372--5384, 2013.

\bibitem{park2008contrast}
G.-H. Park, H.-H. Cho, and M.-R. Choi, ``A contrast enhancement method using dynamic range separate histogram equalization,'' \emph{IEEE Transactions on Consumer Electronics}, vol.~54, no.~4, pp. 1981--1987, 2008.

\bibitem{jobson1997properties}
D.~J. Jobson, Z.-u. Rahman, and G.~A. Woodell, ``Properties and performance of a center/surround retinex,'' \emph{IEEE transactions on image processing}, vol.~6, no.~3, pp. 451--462, 1997.

\bibitem{wang2019underexposed}
R.~Wang, Q.~Zhang, C.-W. Fu, X.~Shen, W.-S. Zheng, and J.~Jia, ``Underexposed photo enhancement using deep illumination estimation,'' in \emph{Proceedings of the IEEE/CVF conference on computer vision and pattern recognition}, 2019, pp. 6849--6857.

\bibitem{jiang2021enlightengan}
Y.~Jiang, X.~Gong, D.~Liu, Y.~Cheng, C.~Fang, X.~Shen, J.~Yang, P.~Zhou, and Z.~Wang, ``Enlightengan: Deep light enhancement without paired supervision,'' \emph{IEEE Transactions on Image Processing}, vol.~30, pp. 2340--2349, 2021.

\bibitem{zhang2022deep}
Z.~Zhang, H.~Zheng, R.~Hong, M.~Xu, S.~Yan, and M.~Wang, ``Deep color consistent network for low-light image enhancement,'' in \emph{Proceedings of the IEEE/CVF conference on computer vision and pattern recognition}, 2022, pp. 1899--1908.

\bibitem{ma2021structure}
Y.~Ma, J.~Liu, Y.~Liu, H.~Fu, Y.~Hu, J.~Cheng, H.~Qi, Y.~Wu, J.~Zhang, and Y.~Zhao, ``Structure and illumination constrained gan for medical image enhancement,'' \emph{IEEE Transactions on Medical Imaging}, vol.~40, no.~12, pp. 3955--3967, 2021.

\bibitem{zhang2016colorful}
R.~Zhang, P.~Isola, and A.~A. Efros, ``Colorful image colorization,'' in \emph{European conference on computer vision}.\hskip 1em plus 0.5em minus 0.4em\relax Springer, 2016, pp. 649--666.

\bibitem{zhang2017real}
R.~Y. Zhang, J.~Y. Zhu, P.~Isola, X.~Geng, A.~S. Lin, T.~Yu, and A.~A. Efros, ``Real-time user-guided image colorization with learned deep priors,'' \emph{ACM Transactions on Graphics}, vol.~36, no.~4, p. 119, 2017.

\bibitem{su2020instance}
J.-W. Su, H.-K. Chu, and J.-B. Huang, ``Instance-aware image colorization,'' in \emph{Proceedings of the IEEE/CVF Conference on Computer Vision and Pattern Recognition}, 2020, pp. 7968--7977.

\bibitem{wu2021towards}
Y.~Wu, X.~Wang, Y.~Li, H.~Zhang, X.~Zhao, and Y.~Shan, ``Towards vivid and diverse image colorization with generative color prior,'' in \emph{Proceedings of the IEEE/CVF International Conference on Computer Vision}, 2021, pp. 14\,377--14\,386.

\bibitem{cao2017unsupervised}
Y.~Cao, Z.~Zhou, W.~Zhang, and Y.~Yu, ``Unsupervised diverse colorization via generative adversarial networks,'' in \emph{Machine Learning and Knowledge Discovery in Databases: European Conference, ECML PKDD 2017, Skopje, Macedonia, September 18--22, 2017, Proceedings, Part I 10}.\hskip 1em plus 0.5em minus 0.4em\relax Springer, 2017, pp. 151--166.

\bibitem{vitoria2020chromagan}
P.~Vitoria, L.~Raad, and C.~Ballester, ``Chromagan: Adversarial picture colorization with semantic class distribution,'' in \emph{Proceedings of the IEEE/CVF Winter Conference on Applications of Computer Vision}, 2020, pp. 2445--2454.

\bibitem{kumar2021colorization}
M.~Kumar, D.~Weissenborn, and N.~Kalchbrenner, ``Colorization transformer,'' \emph{arXiv preprint arXiv:2102.04432}, 2021.

\bibitem{ji2022colorformer}
X.~Ji, B.~Jiang, D.~Luo, G.~Tao, W.~Chu, Z.~Xie, C.~Wang, and Y.~Tai, ``Colorformer: Image colorization via color memory assisted hybrid-attention transformer,'' in \emph{European Conference on Computer Vision}.\hskip 1em plus 0.5em minus 0.4em\relax Springer, 2022, pp. 20--36.

\bibitem{dou2020unpaired}
Q.~Dou, Q.~Liu, P.~A. Heng, and B.~Glocker, ``Unpaired multi-modal segmentation via knowledge distillation,'' \emph{IEEE transactions on medical imaging}, vol.~39, no.~7, pp. 2415--2425, 2020.

\bibitem{qin2021efficient}
D.~Qin, J.-J. Bu, Z.~Liu, X.~Shen, S.~Zhou, J.-J. Gu, Z.-H. Wang, L.~Wu, and H.-F. Dai, ``Efficient medical image segmentation based on knowledge distillation,'' \emph{IEEE Transactions on Medical Imaging}, vol.~40, no.~12, pp. 3820--3831, 2021.

\bibitem{li2020gan}
M.~Li, J.~Lin, Y.~Ding, Z.~Liu, J.-Y. Zhu, and S.~Han, ``Gan compression: Efficient architectures for interactive conditional gans,'' in \emph{Proceedings of the IEEE/CVF conference on computer vision and pattern recognition}, 2020, pp. 5284--5294.

\bibitem{jin2021teachers}
Q.~Jin, J.~Ren, O.~J. Woodford, J.~Wang, G.~Yuan, Y.~Wang, and S.~Tulyakov, ``Teachers do more than teach: Compressing image-to-image models,'' in \emph{Proceedings of the IEEE/CVF Conference on Computer Vision and Pattern Recognition}, 2021, pp. 13\,600--13\,611.

\bibitem{li2020semantic}
Z.~Li, R.~Jiang, and P.~Aarabi, ``Semantic relation preserving knowledge distillation for image-to-image translation,'' in \emph{Computer Vision--ECCV 2020: 16th European Conference, Glasgow, UK, August 23--28, 2020, Proceedings, Part XXVI 16}.\hskip 1em plus 0.5em minus 0.4em\relax Springer, 2020, pp. 648--663.

\bibitem{zhang2022wavelet}
L.~Zhang, X.~Chen, X.~Tu, P.~Wan, N.~Xu, and K.~Ma, ``Wavelet knowledge distillation: Towards efficient image-to-image translation,'' in \emph{Proceedings of the IEEE/CVF Conference on Computer Vision and Pattern Recognition}, 2022, pp. 12\,464--12\,474.

\bibitem{borovec2020anhir}
J.~Borovec, J.~Kybic, I.~Arganda-Carreras, D.~V. Sorokin, G.~Bueno, A.~V. Khvostikov, S.~Bakas, I.~Eric, C.~Chang, S.~Heldmann \emph{et~al.}, ``Anhir: automatic non-rigid histological image registration challenge,'' \emph{IEEE transactions on medical imaging}, vol.~39, no.~10, pp. 3042--3052, 2020.

\bibitem{ding2020multi}
C.~Ding and Z.~Ma, ``Multi-camera color correction via hybrid histogram matching,'' \emph{IEEE Transactions on Circuits and Systems for Video Technology}, vol.~31, no.~9, pp. 3327--3337, 2020.

\bibitem{bihan2019inverse}
B.~Wen, H.~Kadu, and G.-M. Su, ``Inverse luma/chroma mappings with histogram transfer and approximation,'' Apr.~16 2019, uS Patent 10,264,287.

\bibitem{mao2017least}
X.~Mao, Q.~Li, H.~Xie, R.~Y. Lau, Z.~Wang, and S.~Paul~Smolley, ``Least squares generative adversarial networks,'' in \emph{Proceedings of the IEEE international conference on computer vision}, 2017, pp. 2794--2802.

\bibitem{hu2021bidirectional}
S.~Hu, B.~Lei, S.~Wang, Y.~Wang, Z.~Feng, and Y.~Shen, ``Bidirectional mapping generative adversarial networks for brain mr to pet synthesis,'' \emph{IEEE Transactions on Medical Imaging}, vol.~41, no.~1, pp. 145--157, 2021.

\bibitem{colleoni2022ssis}
E.~Colleoni, D.~Psychogyios, B.~Van~Amsterdam, F.~Vasconcelos, and D.~Stoyanov, ``Ssis-seg: Simulation-supervised image synthesis for surgical instrument segmentation,'' \emph{IEEE Transactions on Medical Imaging}, vol.~41, no.~11, pp. 3074--3086, 2022.

\bibitem{kong2021breaking}
L.~Kong, C.~Lian, D.~Huang, Y.~Hu, Q.~Zhou \emph{et~al.}, ``Breaking the dilemma of medical image-to-image translation,'' \emph{Advances in Neural Information Processing Systems}, vol.~34, pp. 1964--1978, 2021.

\bibitem{aly2005image}
H.~A. Aly and E.~Dubois, ``Image up-sampling using total-variation regularization with a new observation model,'' \emph{IEEE Transactions on Image Processing}, vol.~14, no.~10, pp. 1647--1659, 2005.

\bibitem{isola2017image}
P.~Isola, J.-Y. Zhu, T.~Zhou, and A.~A. Efros, ``Image-to-image translation with conditional adversarial networks,'' in \emph{Proceedings of the IEEE conference on computer vision and pattern recognition}, 2017, pp. 1125--1134.

\bibitem{heusel2017gans}
M.~Heusel, H.~Ramsauer, T.~Unterthiner, B.~Nessler, and S.~Hochreiter, ``Gans trained by a two time-scale update rule converge to a local nash equilibrium,'' \emph{Advances in neural information processing systems}, vol.~30, 2017.

\bibitem{bińkowski2018demystifying}
M.~Bińkowski, D.~J. Sutherland, M.~Arbel, and A.~Gretton, ``Demystifying {MMD} {GAN}s,'' in \emph{International Conference on Learning Representations}, 2018.

\bibitem{mittal2012making}
A.~Mittal, R.~Soundararajan, and A.~C. Bovik, ``Making a “completely blind” image quality analyzer,'' \emph{IEEE Signal processing letters}, vol.~20, no.~3, pp. 209--212, 2012.

\bibitem{xie2022unsupervised}
S.~Xie, Q.~Ho, and K.~Zhang, ``Unsupervised image-to-image translation with density changing regularization,'' \emph{Advances in Neural Information Processing Systems}, vol.~35, pp. 28\,545--28\,558, 2022.

\bibitem{xie2023unpaired}
S.~Xie, Y.~Xu, M.~Gong, and K.~Zhang, ``Unpaired image-to-image translation with shortest path regularization,'' in \emph{Proceedings of the IEEE/CVF Conference on Computer Vision and Pattern Recognition}, 2023, pp. 10\,177--10\,187.

\bibitem{zhang2018unreasonable}
R.~Zhang, P.~Isola, A.~A. Efros, E.~Shechtman, and O.~Wang, ``The unreasonable effectiveness of deep features as a perceptual metric,'' in \emph{Proceedings of the IEEE conference on computer vision and pattern recognition}, 2018, pp. 586--595.

\bibitem{han2023multilevel}
M.~Han, M.~Shao, L.~Meng, Y.~Liu, and Y.~Qiao, ``Multilevel contrast strategy for unpaired image-to-image translation,'' \emph{Journal of Electronic Imaging}, vol.~32, no.~6, pp. 063\,030--063\,030, 2023.

\bibitem{zhan2022modulated}
F.~Zhan, J.~Zhang, Y.~Yu, R.~Wu, and S.~Lu, ``Modulated contrast for versatile image synthesis,'' in \emph{Proceedings of the IEEE/CVF Conference on Computer Vision and Pattern Recognition}, 2022, pp. 18\,280--18\,290.

\bibitem{hu2022qs}
X.~Hu, X.~Zhou, Q.~Huang, Z.~Shi, L.~Sun, and Q.~Li, ``Qs-attn: Query-selected attention for contrastive learning in i2i translation,'' in \emph{Proceedings of the IEEE/CVF Conference on Computer Vision and Pattern Recognition}, 2022, pp. 18\,291--18\,300.

\bibitem{liang2021swinir}
J.~Liang, J.~Cao, G.~Sun, K.~Zhang, L.~Van~Gool, and R.~Timofte, ``Swinir: Image restoration using swin transformer,'' in \emph{Proceedings of the IEEE/CVF international conference on computer vision}, 2021, pp. 1833--1844.

\bibitem{saharia2022palette}
C.~Saharia, W.~Chan, H.~Chang, C.~Lee, J.~Ho, T.~Salimans, D.~Fleet, and M.~Norouzi, ``Palette: Image-to-image diffusion models,'' in \emph{ACM SIGGRAPH 2022 Conference Proceedings}, 2022, pp. 1--10.

\bibitem{DINO}
\BIBentryALTinterwordspacing
K.~Vougioukas, S.~Petridis, and M.~Pantic, ``{DINO:} {A} conditional energy-based {GAN} for domain translation,'' in \emph{9th International Conference on Learning Representations, {ICLR} 2021, Virtual Event, Austria, May 3-7, 2021}.\hskip 1em plus 0.5em minus 0.4em\relax OpenReview.net, 2021. [Online]. Available: \url{https://openreview.net/forum?id=WAISmwsqDsb}
\BIBentrySTDinterwordspacing

\bibitem{cohen2018distribution}
J.~P. Cohen, M.~Luck, and S.~Honari, ``Distribution matching losses can hallucinate features in medical image translation,'' in \emph{Medical Image Computing and Computer Assisted Intervention--MICCAI 2018: 21st International Conference, Granada, Spain, September 16-20, 2018, Proceedings, Part I}.\hskip 1em plus 0.5em minus 0.4em\relax Springer, 2018, pp. 529--536.

\bibitem{tsutsui2024wbcatt}
S.~Tsutsui, W.~Pang, and B.~Wen, ``Wbcatt: A white blood cell dataset annotated with detailed morphological attributes,'' \emph{Advances in Neural Information Processing Systems}, vol.~36, 2024.

\end{thebibliography}

\end{document}